\documentclass[runningheads]{llncs}
\usepackage{graphicx}

\usepackage{tikz}
\usepackage{comment}
\usepackage{amsmath,amssymb} 
\usepackage{color}
\usepackage{subcaption}

\usepackage[accsupp]{axessibility}  

\usepackage[width=122mm,left=12mm,paperwidth=146mm,height=193mm,top=12mm,paperheight=217mm]{geometry}
\usepackage{booktabs}

\begin{document}
\pagestyle{headings}
\mainmatter
\def\ECCVSubNumber{100}  

\title{BON: An extended public domain dataset for human activity recognition } 


\titlerunning{BON: An extended public domain dataset for human activity recognition} 
%
\author{Girmaw Abebe Tadesse \and
Oliver  Bent \and  Komminist Weldemariam \and  Md. Abrar Istiak \and Taufiq Hasan \and Andrea Cavallaro}
%
%
\authorrunning{G.A. Tadesse et al.}
%
\maketitle

\begin{abstract}
Body-worn \textit{first-person vision} (FPV) camera enables to extract a rich source of information on  the environment from the subject's viewpoint. However, the research progress in wearable camera-based egocentric office activity understanding is slow compared to other activity environments (e.g., kitchen and outdoor ambulatory), mainly due to the lack of adequate datasets to train more sophisticated  (e.g., deep learning) models for human activity recognition in office environments.
This paper provides details of a large and publicly available office activity dataset (BON) collected in different office settings across  three  geographical locations: Barcelona (Spain), Oxford (UK) and Nairobi (Kenya), using a chest-mounted GoPro Hero camera. 
The BON dataset contains eighteen common office activities that can be categorised into person-to-person interactions (e.g., Chat with colleagues), person-to-object (e.g., Writing on a whiteboard), and proprioceptive (e.g., Walking). Annotation is provided for each segment of video with 5-seconds duration. Generally, BON contains 25 subjects and 2639 total segments. In order to facilitate further research in the sub-domain, we have also provided  results that could be used as baselines for future studies. 

\end{abstract}

\section{Introduction}

Human activity recognition (HAR) is one of the well-explored domains but still requires further studies due to its challenges, mainly associated with the variability in: sensors (e.g., inertial measurement units and camera), subjects (e.g., gender, age and gait), activities (e.g., ambulatory and interactive) and environments (e.g., indoor and outdoor). HAR plays a crucial role in the fields of human-computer interactions, assistive computer vision and robotics. Particularly, activity recognition in workplaces (aka office activity recognition)  includes personal activity tracking of a subject that could be utilised to maximise productivity and safety.

Various modalities have been employed to encode information for the task of human activity recognition. These include inertial measurement units (e.g., accelerometer)~\cite{kwapisz2011activity,wan2020deep,lu2020efficient,abebe2017inertial}, radar~\cite{zhu2018indoor,wang2015understanding}, WiFi~\cite{wang2015understanding} signals, RGB
depth video~\cite{ni2011rgbd}, infrared~\cite{gao2016infar,jiang2017learning}, skeleton~\cite{liu2017pku} and point cloud~\cite{cheng2016orthogonal} data.

\begin{figure}[t]
    \centering
    \includegraphics[width=0.8\linewidth]{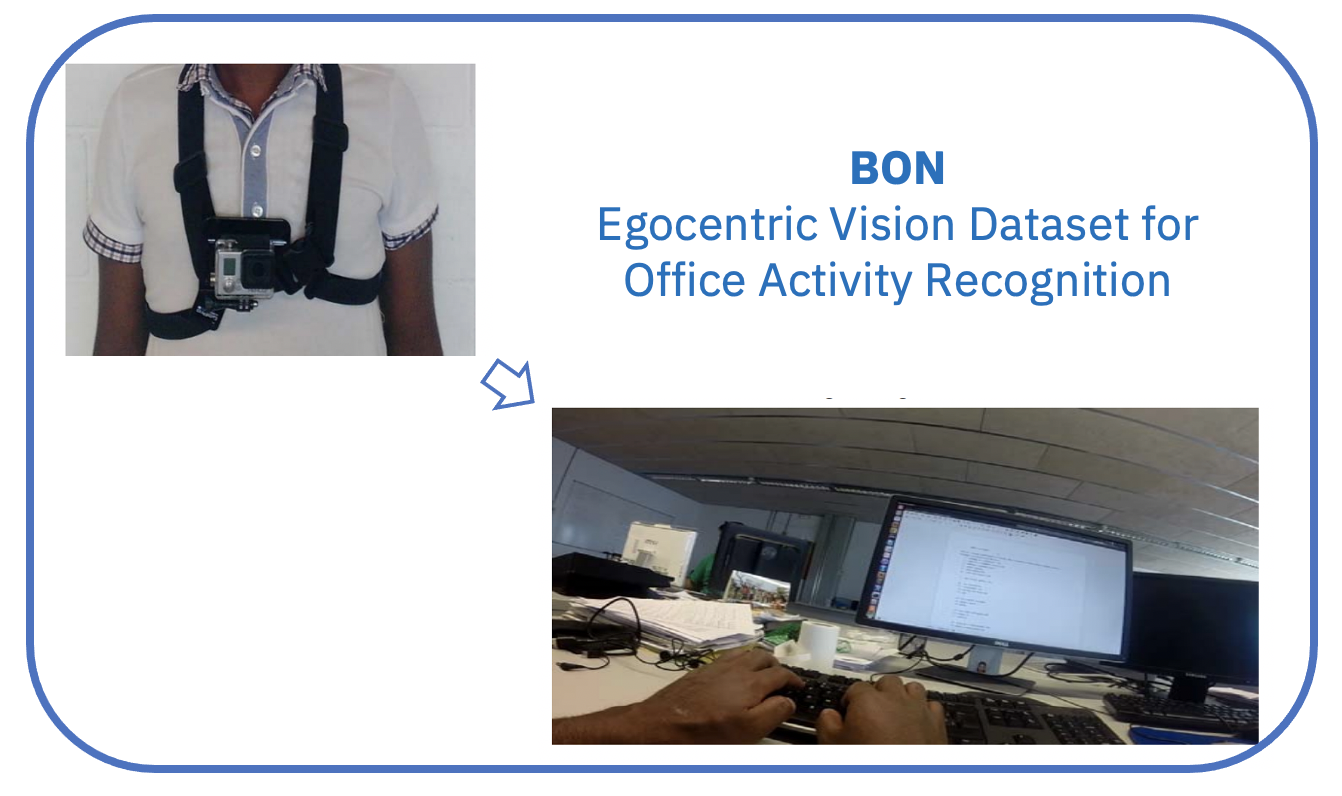}
    \caption{Chest-mounted wearable camera setting, used for collecting the BON dataset, and an example of a first-person video (FPV) frame from the dataset.}
    \label{fig:bon_overview}
\end{figure}

The development of cheaper and easy-to-use cameras has made huge progress in collecting more image/video data, which provide richer information in both spatial and temporal dimensions while privacy and environmental sensitivity could be an issue. Depending on the topology between camera used and active subject under consideration (for which the information is collected for), visual data could be collected from two perspectives: traditional (third-person vision) and egocentric (first-person vision). In a complex and highly dynamic environment where the density of objects is high, a third-person camera provides a global view of
the high-level attributes in a scene and preferred to understand group-based human activities but stills susceptible to privacy issues as the active user is clearly seen in this setting. On the other hand, first-person camera can capture finer details about objects
and people with superior level of granularity from the user's field of view while privacy is relatively protected as the active user will not be seen in the video collected (see Fig.~\ref{fig:bon_overview}).


Understanding of egocentric human activities is crucial across different application domains. A significant research progress has been made in recent years to identify kitchen~\cite{fathi2012learning,Damen2018EPICKITCHENS} or outdoor~\cite{abebe2017long,furnari2016recognizing,abebe2016robust,abebe2017hierarchical} activities. 
 However, the research progress in wearable camera-based egocentric office activity understanding is still minimal, partly due to the lack of adequate datasets to train more sophisticated  (e.g., deep learning)~\cite{ogaki2012coupling,abebe2018first}. Wojek et al.~\cite{wojek2006activity} reported one  of the earlier works in office activity understanding that employed fusion of audio and (third-person) video data using hidden Markov models (HMMs).
The set of studied activities  in this work includes \textit{meetings, paperwork, phone calls}, and \textit{nobody present}. Ying et al.~\cite{wang2007abnormal} detected abnormal activities in office setting by utilising Radon transform on binary human silhouette. This has been later extended to encompass twelve office activities using hand posture cues in~\cite{paulson2008office}. Tadesse et al.~\cite{abebe2018first} extended the number office activities to eighteen but using FPV videos collected from 12 subjects in unique office setting. Recognition of office activities have also been reported using sensors other than wearable camera, such as accelerometer~\cite{cha2018towards}, passive infrared, pressure and acoustic sensors~\cite{nguyen2014ontology}, gyroscope~\cite{mekruksavanich2018smartwatch}.

In this paper we present BON - a large and publicly available FPV dataset of office activities, which was extended from ~\cite{abebe2018first} by diversifying the data collection settings. BON is available at ~\cite{girmaw2021bon} and it consists of  $2639$ video segments.  First, we present the data collection and annotation set-up followed by critical challenges associated with encoding human activities from first-person videos of users (Section~\ref{data_descr}).  The BON dataset was previously utilized for Video and Image Processing (VIP) Cup 2019 challenge~\cite{tadesse2020privacy}, and baseline results from the challenge are also provided  in Section~\ref{experim} so that future works could compare against these baselines.  Finally, Section~\ref{conclu}  concludes the paper. 

\section{BON: Large scale office activities dataset}
 \label{data_descr}
\textit{BON} represents a new FPV dataset of office activities collected in \textbf{B}arcelona (Spain), \textbf{O}xford (UK) and \textbf{N}airobi (Kenya)  using a GoPro Hero wearable camera~\cite{girmaw2021bon}. BON is extended from ~\cite{abebe2018first} by diversifying the collection settings across different countries along with increased subjects' participation. Multiple office settings are included in BON dataset, which will help to develop generalisable HAR models that are robust against variations.  In this Section, we describe the details of BON including data collection setting, definition of activities, segments stratification (e.g., across locations). We have also highlighted technical challenges available in the BON dataset (e.g., specific inter-class similarity and intra-class variations in office settings).

 \subsection{Data collection}
\begin{figure*}
\subfloat[Clean]{
  \includegraphics[height=15mm,width=30mm]{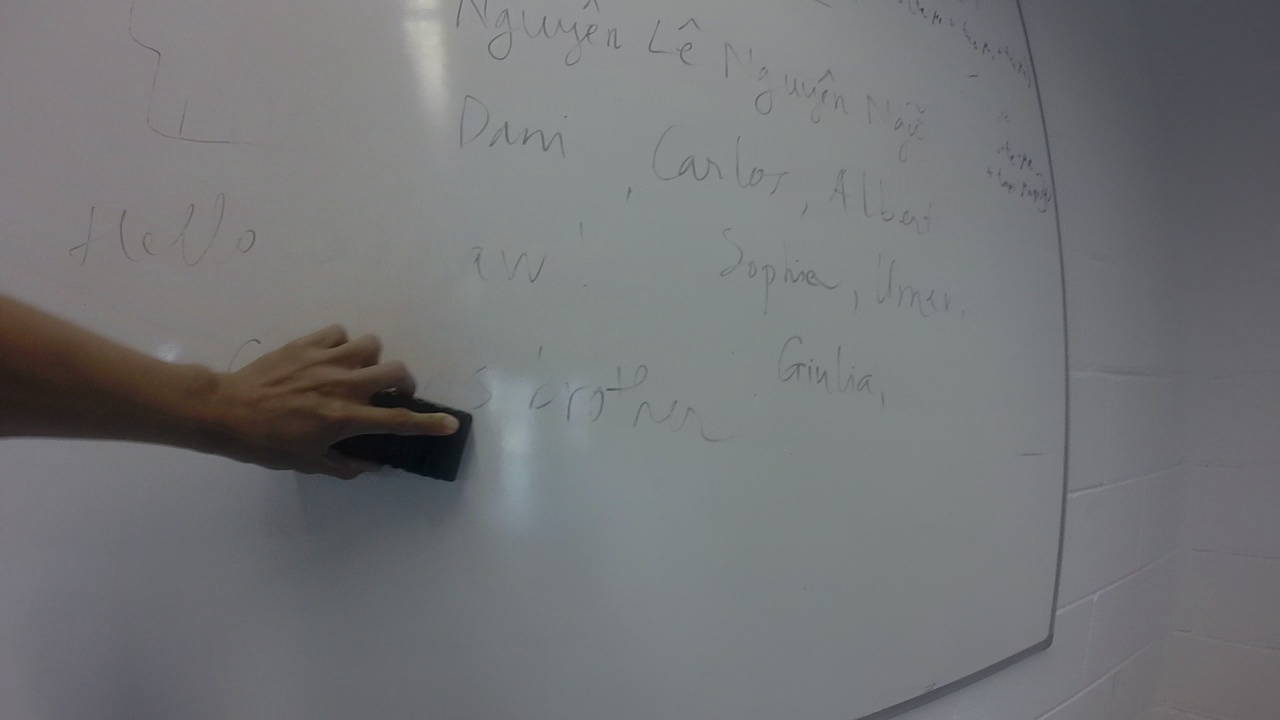}
}
\subfloat[Drink]{
  \includegraphics[height=15mm,width=30mm]{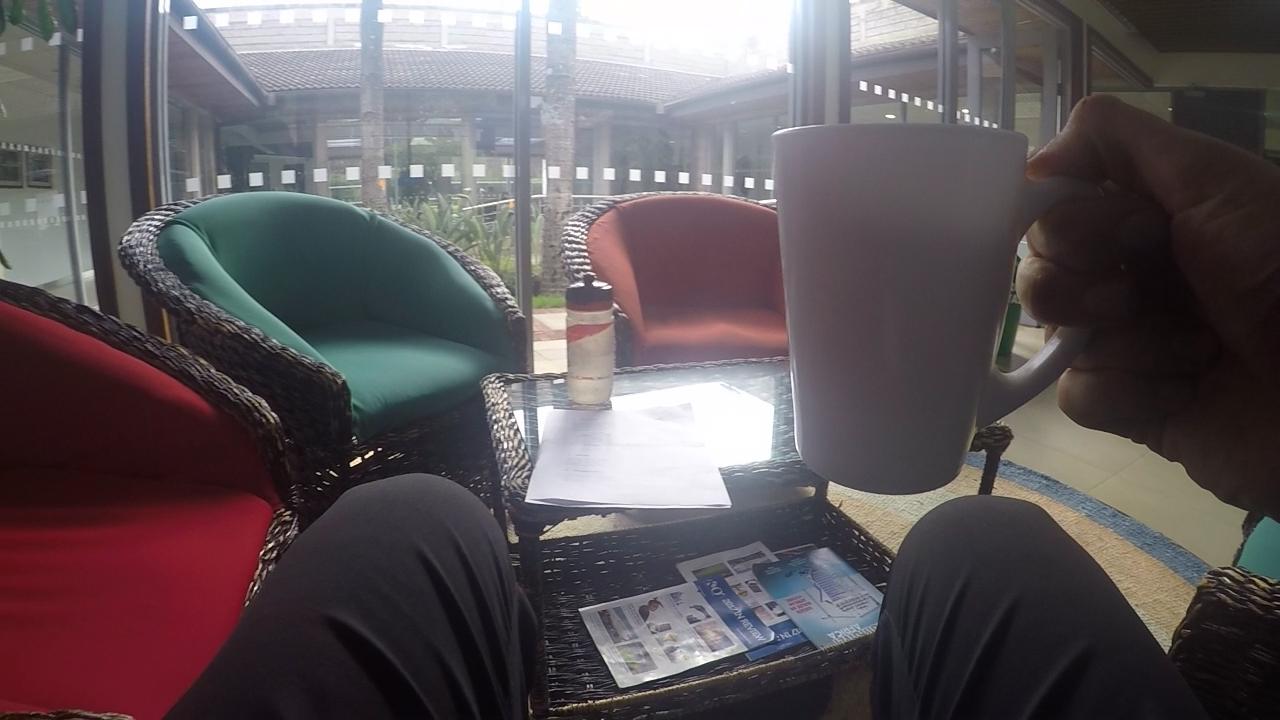}
}
\subfloat[Dry]{
  \includegraphics[height=15mm,width=30mm]{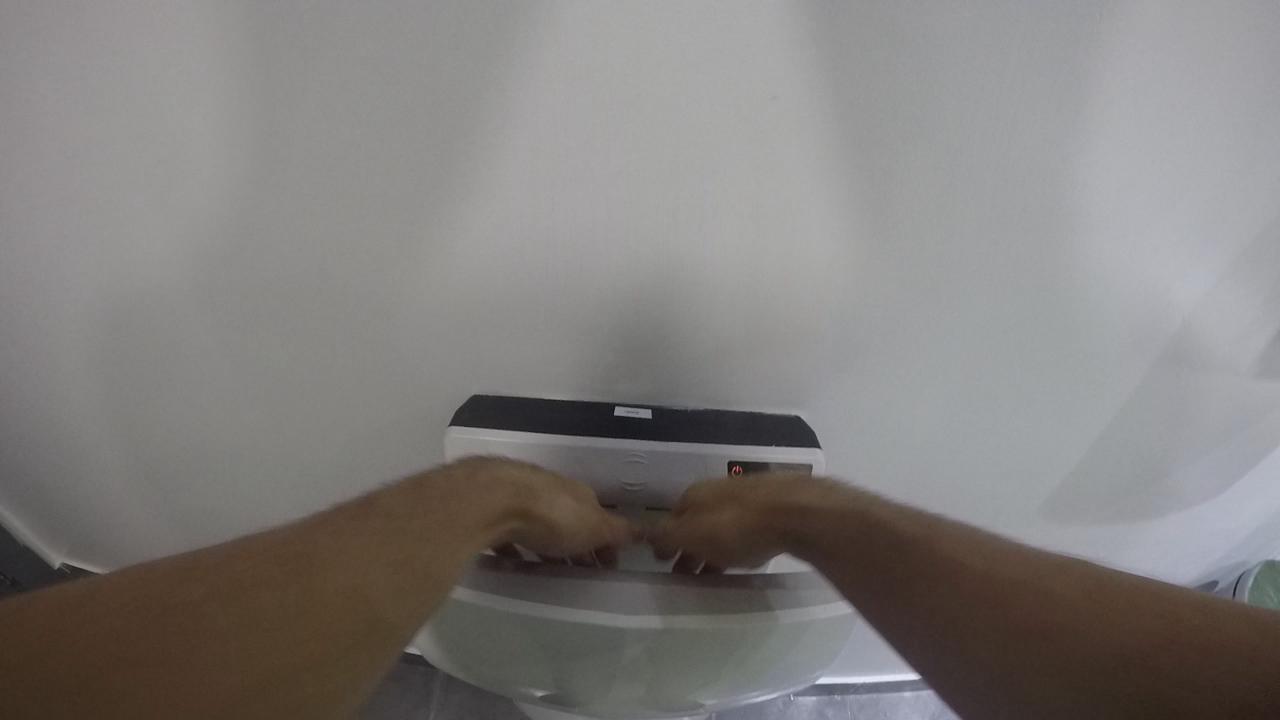}
}
\subfloat[Machine]{
  \includegraphics[height=15mm,width=30mm]{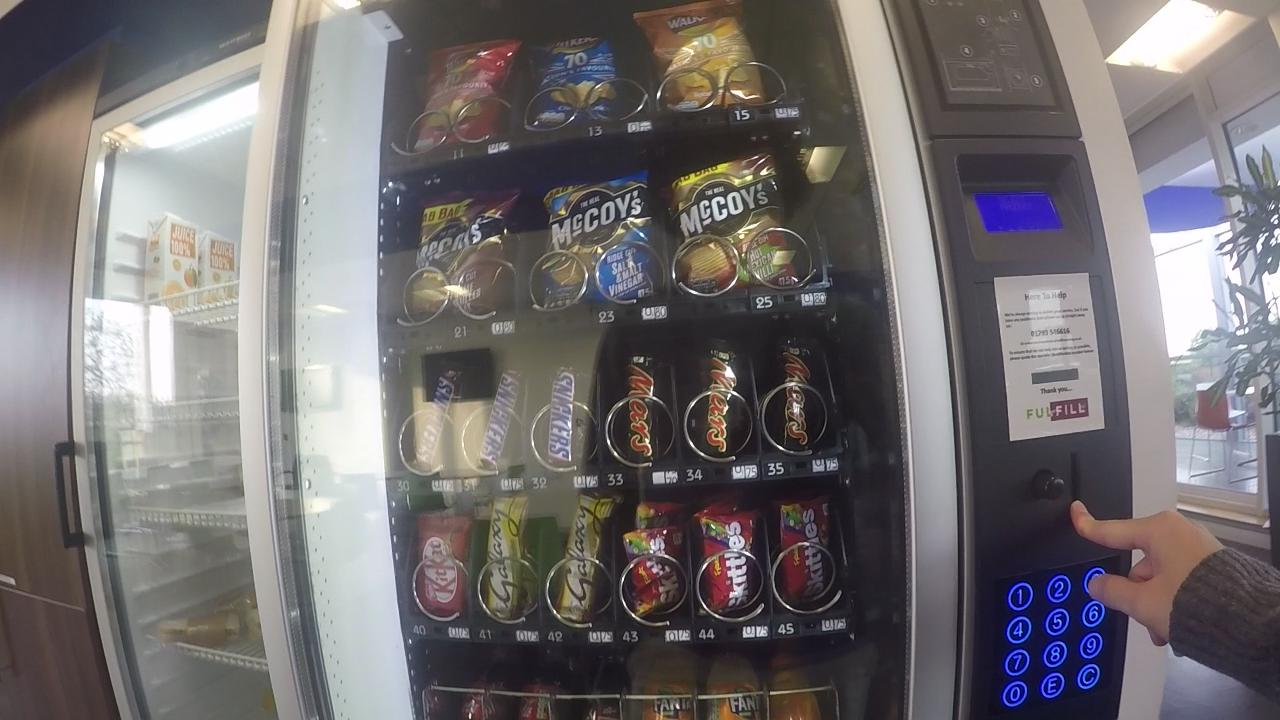}
}

\subfloat[Microwave]{
  \includegraphics[height=17mm,width=30mm]{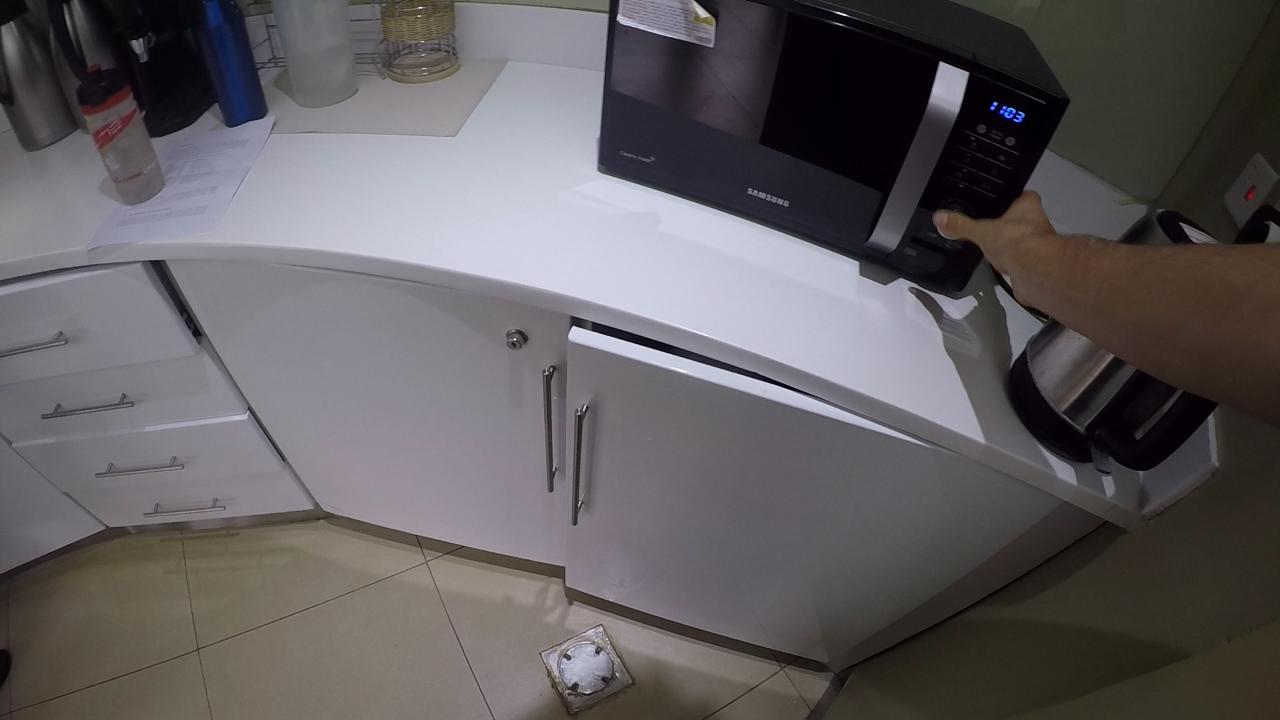}
}
\subfloat[Mobile]{
  \includegraphics[height=17mm,width=30mm]{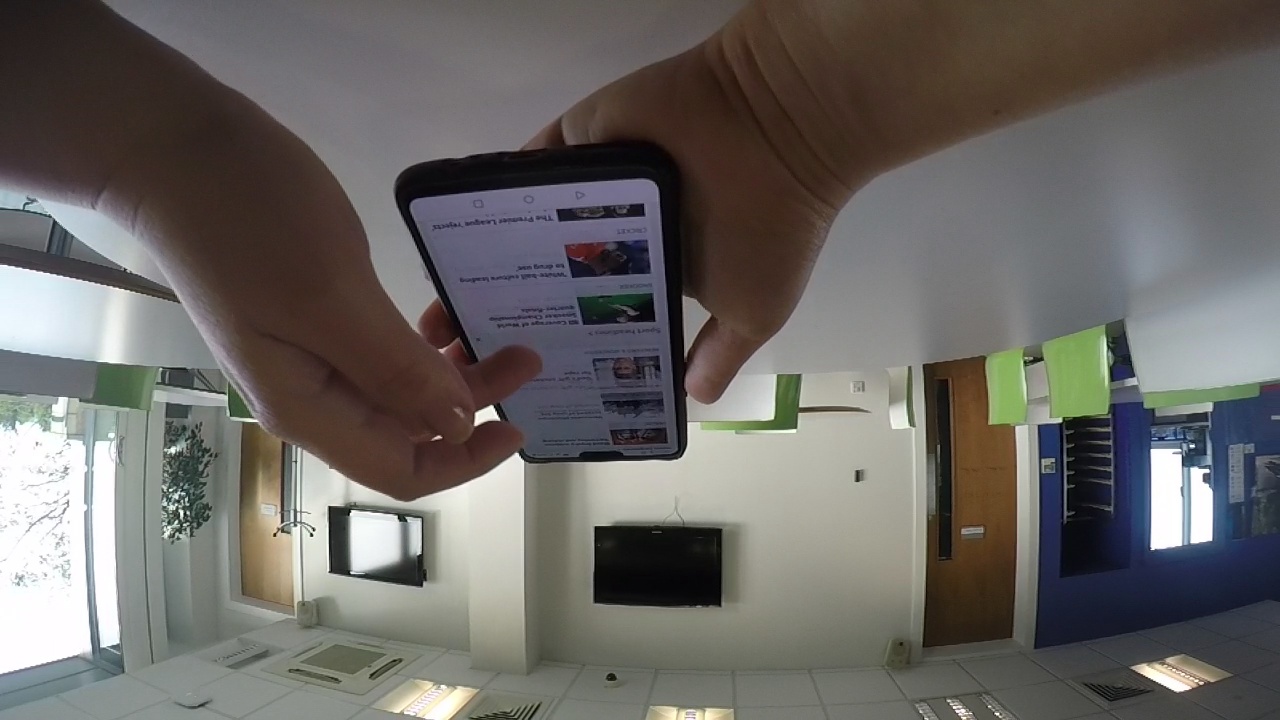}
}
\subfloat[Paper]{
  \includegraphics[height=17mm,width=30mm]{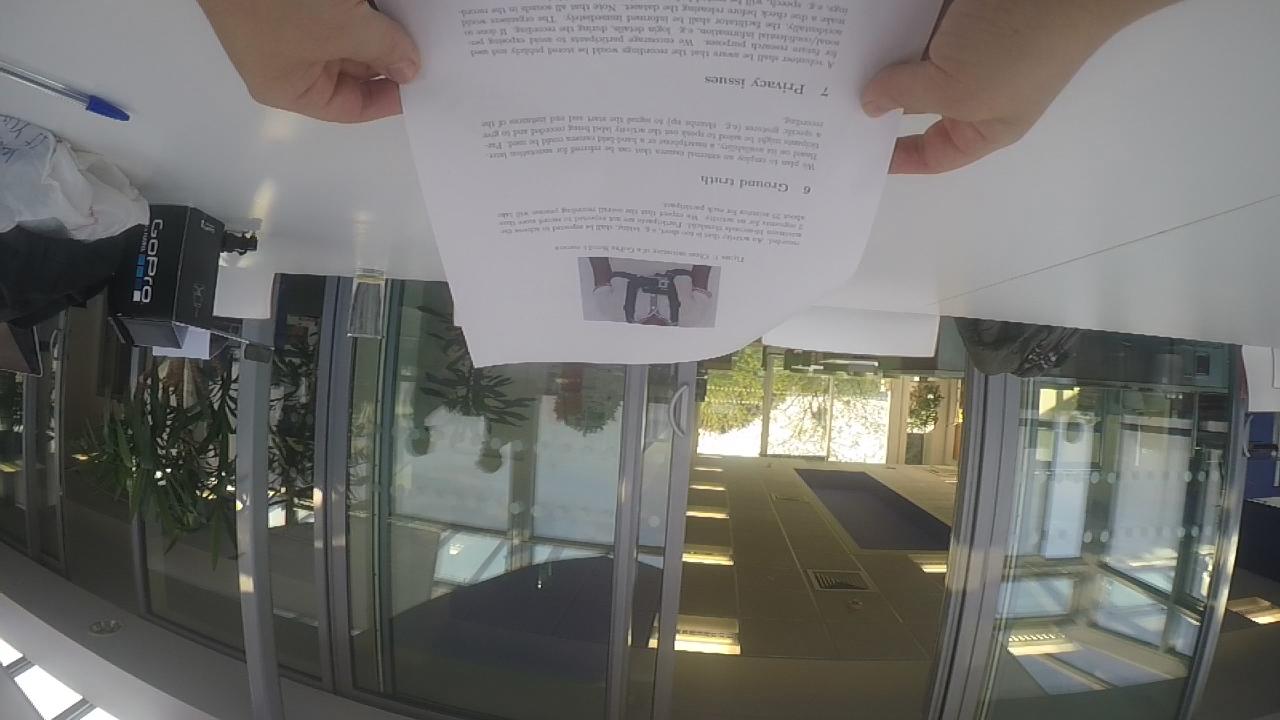}
}
\subfloat[Print]{
  \includegraphics[height=17mm,width=30mm]{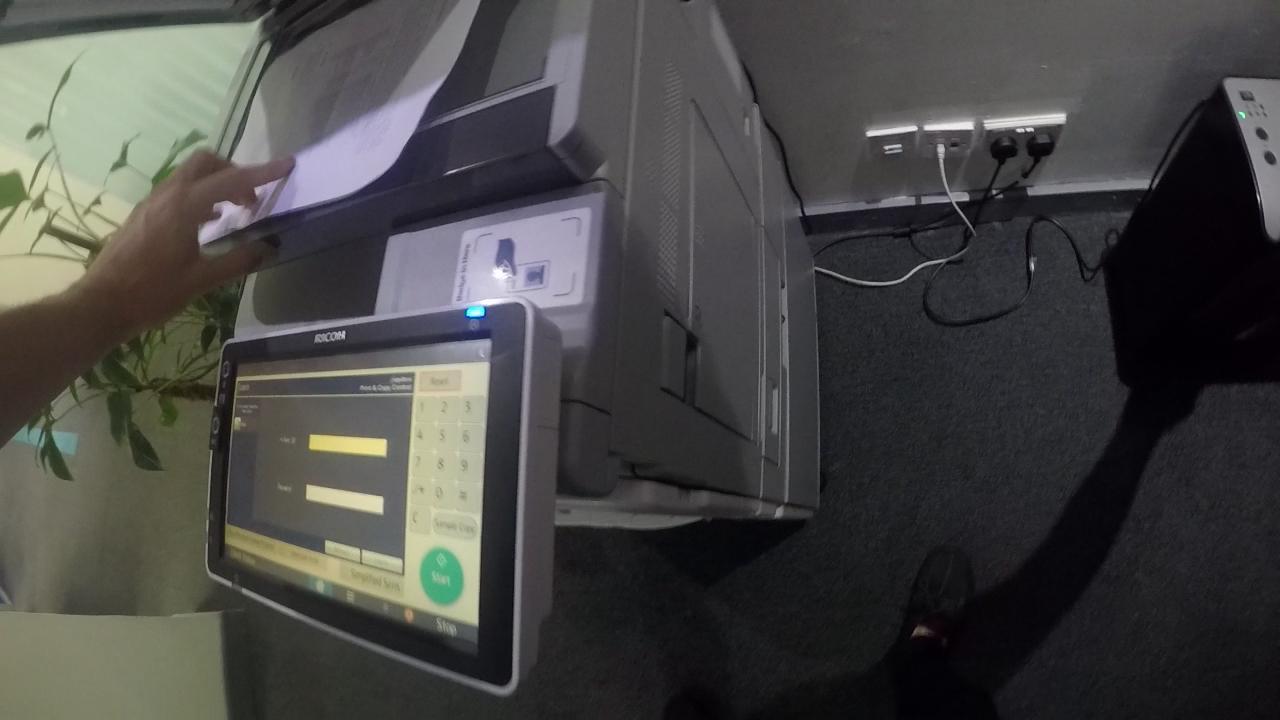}
}

\subfloat[Read]{
  \includegraphics[height=17mm,width=30mm]{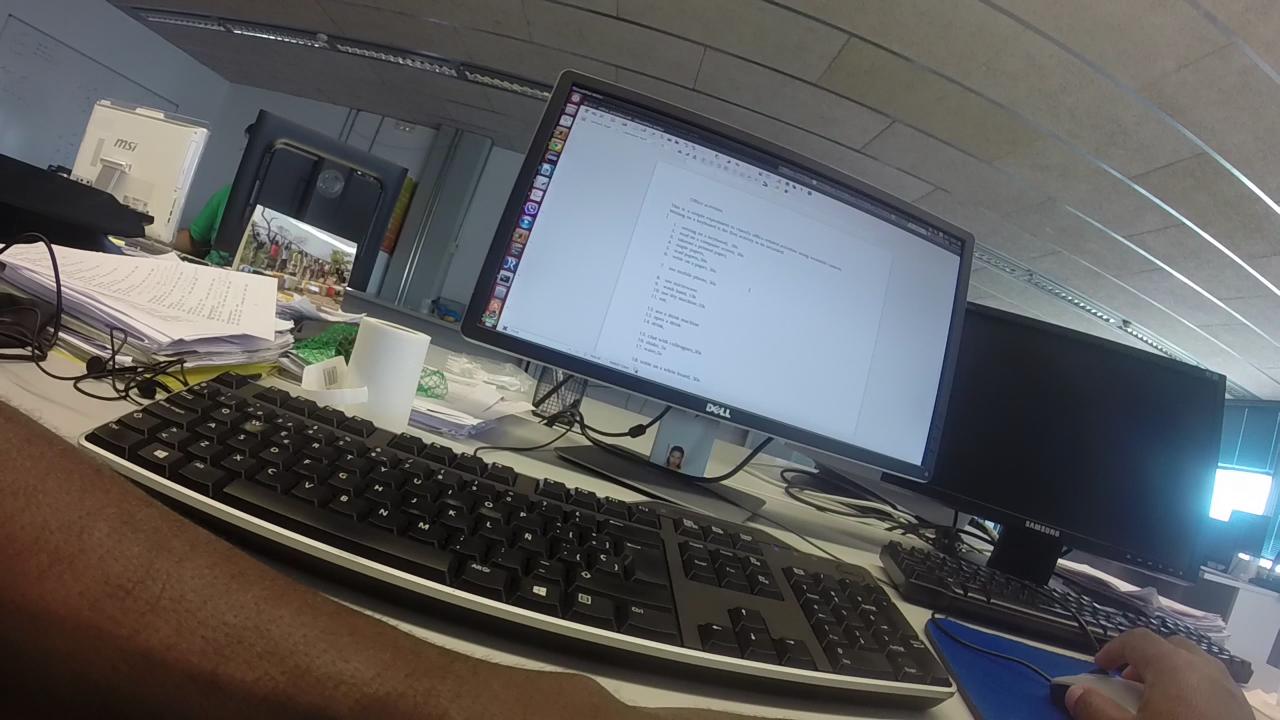}
}
\subfloat[Staple]{
  \includegraphics[height=17mm,width=30mm]{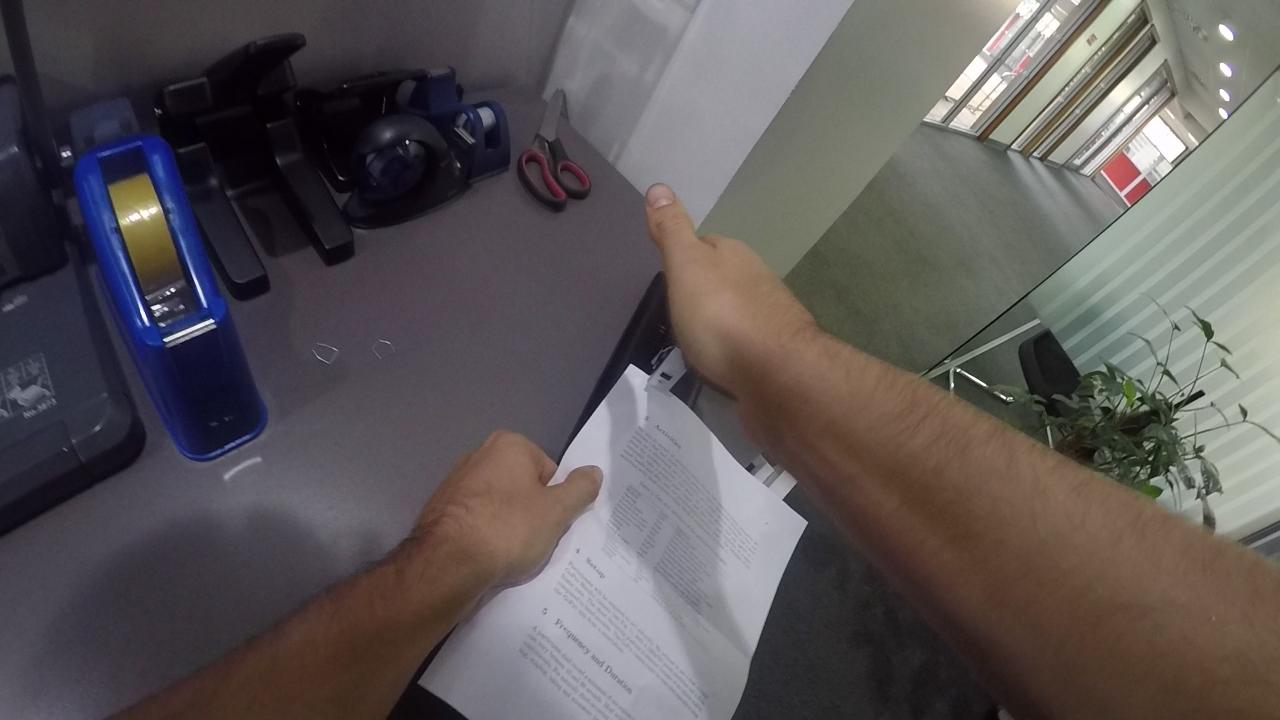}
}
\subfloat[Take]{
  \includegraphics[height=17mm,width=30mm]{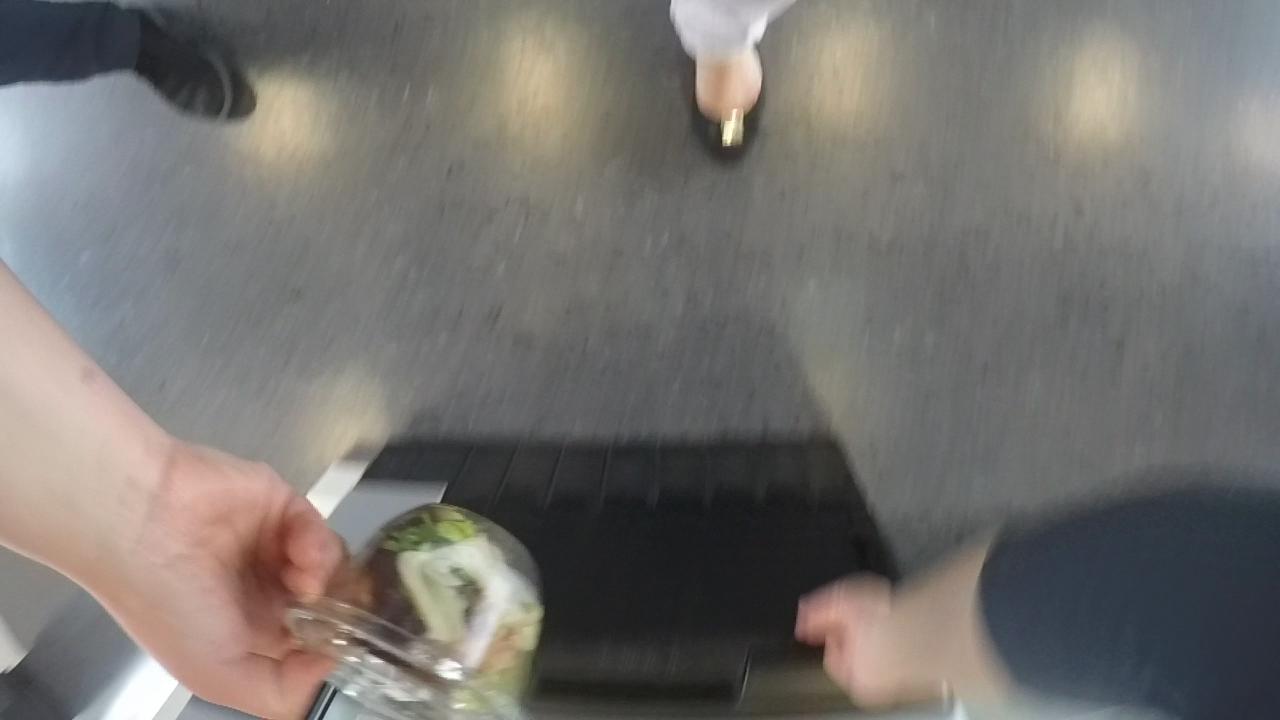}
}
\subfloat[Typeset]{
  \includegraphics[height=17mm,width=30mm]{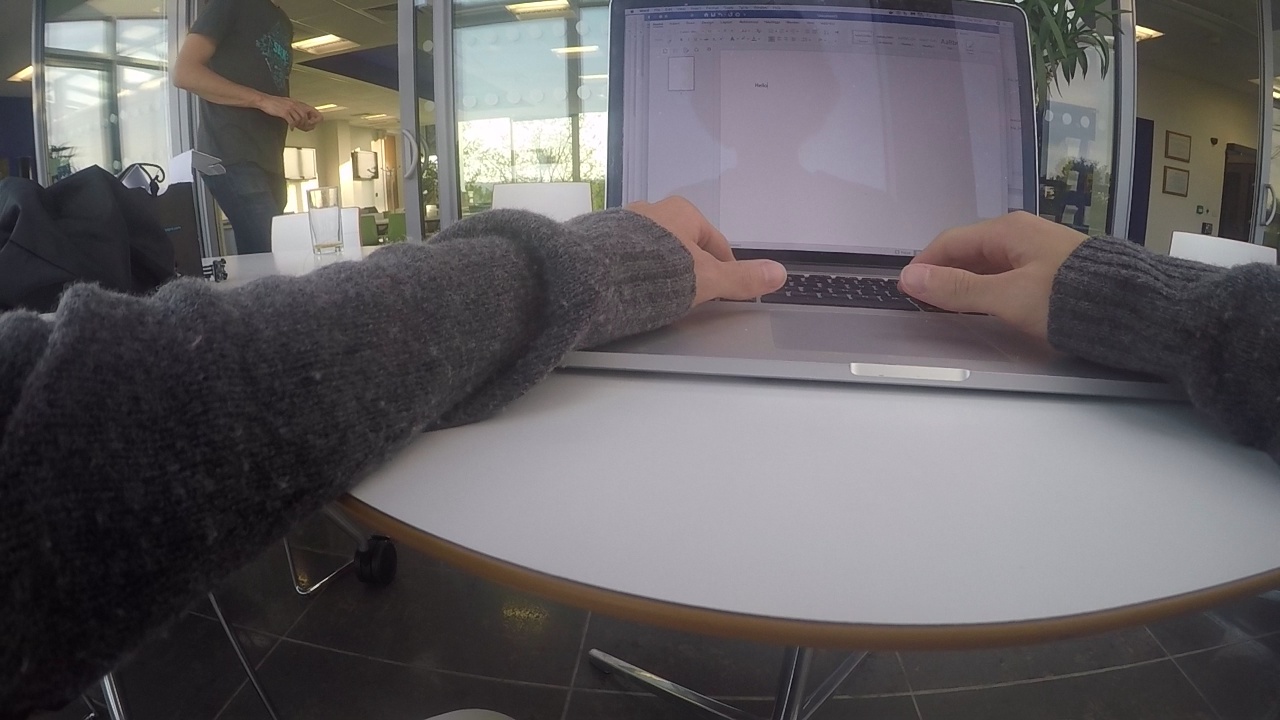}
}

\subfloat[Wash]{
  \includegraphics[height=17mm,width=30mm]{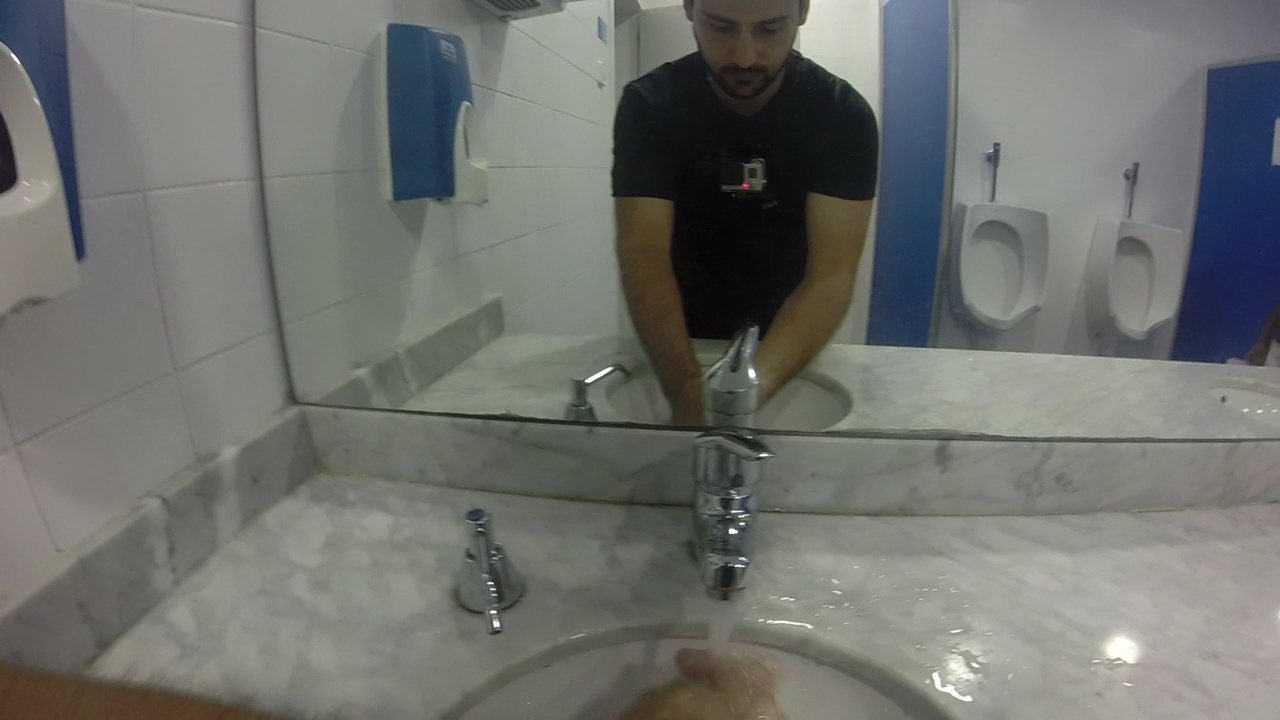}
}
\subfloat[Whiteboard]{
  \includegraphics[height=17mm,width=30mm]{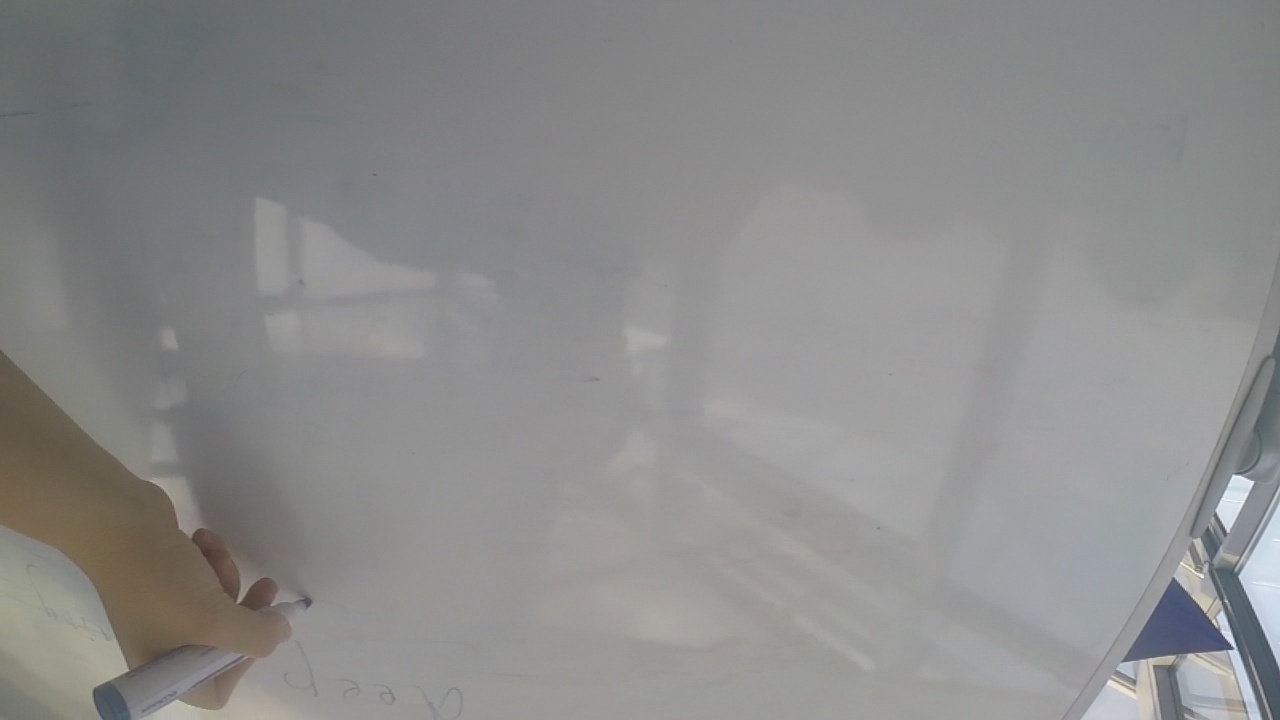}
}
\subfloat[Write]{
  \includegraphics[height=17mm,width=30mm]{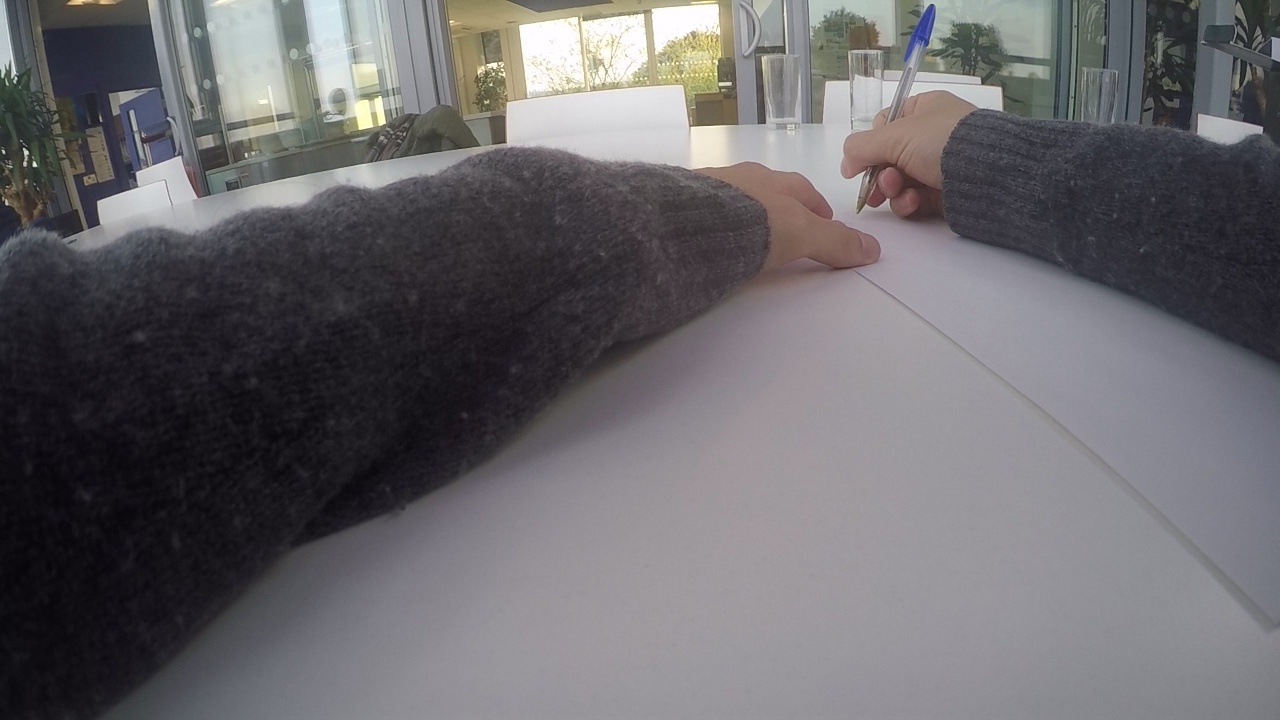}
}
\subfloat[Walk]{
  \includegraphics[height=17mm,width=30mm]{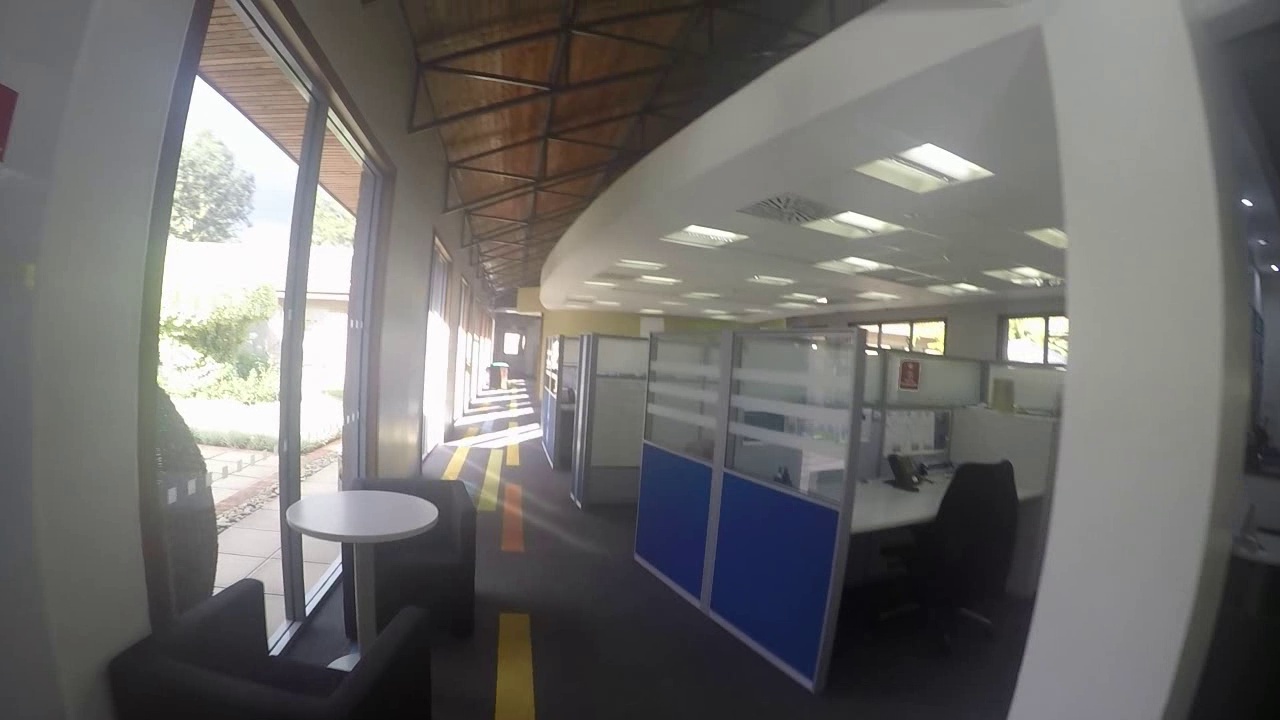}
}

\centering
\subfloat[Chat]{
  \includegraphics[height=17mm,width=30mm]{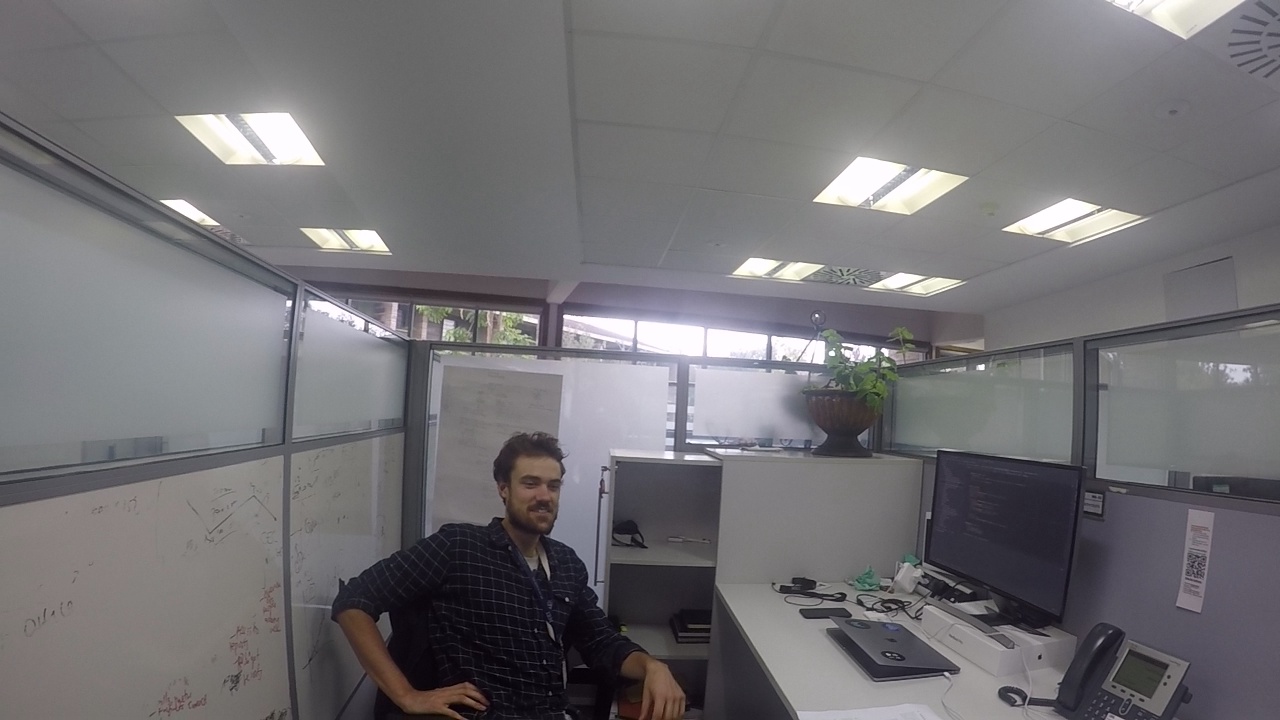}
}
\subfloat[Shake]{
  \includegraphics[height=17mm,width=30mm]{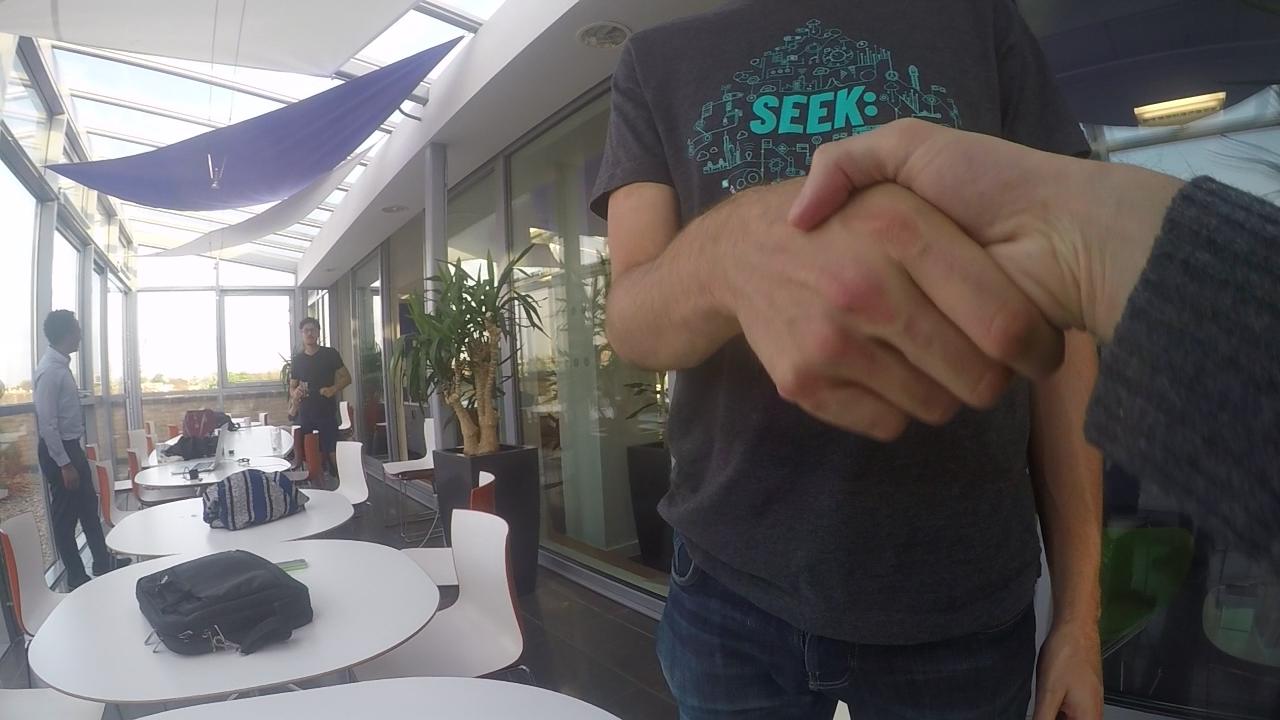}
}

\caption{Key frames from sampled videos of $18$ activities in the BON dataset that can be grouped into three categories. (a) - (o) \textit{person-to-object interaction} category, (p)  \textit{proprioceptive} category, and (q) - (r) \textit{person-to- person} category.}
\label{samplefig}
\end{figure*}

 A chest-mounted GoPro Hero3+ camera was used to collect video segments of the  BON dataset with a resolution of $1280\times 720$ and a rate of  $30$ frames per second.
 Participants were instructed to record themselves (i.e., to press the recording button) while they were performing common office activities. In total there are $18$ activities in BON, which can be categorized into three categories: \textit{Person-to-object interactions}, \textit{Person-to-person interactions}, and  \textit{proprioceptive} activity. A descriptive definition of all activities is given in Table~\ref{table:active_def}. Randomly selected samples from each activity are also shown in Fig.~\ref{samplefig}.

 A total of $25$ subjects were participated in the collection of BON: $12$ subjects in Barcelona, $7$ subjects in Oxford, and $6$ subjects in Nairobi sub-datasets.  A quantitative description is given in Table~\ref{table:locwiseData}.   Each subject recorded a continuous video sequence. Then a pre-processing is  employed to segment a continuous recording of activity into a video sample of 5-seconds duration. 
This results in a total of $2639$ video segments, and the distribution of these segments across the three locations is presented in Table~\ref{table:active_distribution}. When stratified across locations, $1308$ segments were collected in Oxford, $464$ segments in  Barcelona  and $867$ segments Nairobi. Among activities \textit{shake,  staple}, and \textit{take} represented the minority groups.

\begin{table}[t]
\caption{Definition of activities in the BON dataset}\label{table:active_def}
\centering
 \resizebox{0.80\linewidth}{!}{
\begin{tabular}{ll}
\toprule
\textbf{Label} & \textbf{Definition}\\ \midrule
Chat & Chatting with other person\\
Clean & Cleaning the whiteboard using a duster or an ink remover\\
Drink & Drinking from a can or a cup or bottle or glass\\
Dry & Drying hand using electric hand dryer\\
Machine & Placing an order in a vending machine\\
Microwave & Heating food using a microwave\\
Mobile & Scrolling mobile screen\\
Paper & Reading from a paper\\
Print & Taking out printed paper from a printer or xerox machine\\
Read & Reading from a computer screen\\
Shake & Shaking hand with another person\\
Staple & Stapling papers using a stapler\\
Take & Taking out a bottle or a can or a container from a vending machine\\
Typeset & Typing computer keyboard\\
Walk & Walking naturally\\
Wash & Washing hand in a sink\\
Whiteboard & Writing on whiteboard using a marker\\
Write & Writing on paper using a pen or a pencil\\
\bottomrule
\end{tabular}
}
\end{table}

\begin{table}[t]
\centering
\caption{Stratification of collected video segments across locations.}\label{table:locwiseData}
 \resizebox{0.7\linewidth}{!}{
\begin{tabular}{lc|cc|c}
\toprule
&&\multicolumn{2}{c|}{Sex}& \\
Location    & Num. of Subjects  & Male & Female & Num. of Segments \\ [0.4ex] 
\midrule
Barcelona  & $12$ & $9$ & $3$ & $464$ \\[0.1ex]
Oxford  & $7$ & 5 & 2 & $1308$ \\[0.1ex]
Nairobi  & $6$ & 4 & 2 & $867$ \\[0.1ex]
\midrule
Total  & $25$ & 18 & 7 & $\textbf{2639}$ \\[0.1ex]
\bottomrule
\end{tabular}
}

\end{table}

\begin{table}[h]
\caption{Distribution of office activities across the three geographical locations in the BON dataset.}\label{table:active_distribution}
\centering
 \resizebox{0.5\linewidth}{!}{
\begin{tabular}{lccc|c}
\toprule
& \multicolumn{3}{c}{Location} \\
Label & Barcelona & Nairobi & Oxford & Total \\ \midrule
Chat & 30   & 61    &  99 &  190 \\[0.1ex]
Clean & 19  &  44   & 54 &  117 \\[0.1ex]
Drink & 22   &  53    & 69 & 144  \\[0.1ex]
Dry & 16  &   24  & 61 &  101  \\[0.1ex]
Machine &  21 &  42   &  76 & 139 \\[0.1ex]
Microwave & 27  &  60   & 93 & 180  \\[0.1ex]
Mobile &  13 &   46  & 74 & 133  \\[0.1ex]
Paper & 22  &   45  & 92 & 159  \\[0.1ex]
Print & 17   &    51  & 53 & 121  \\[0.1ex]
Read &  20 &   52  & 79 &  151 \\[0.1ex]
Shake &  17 &  18   & 34 &  69 \\[0.1ex]
Staple &  16 &  30   & 49 & 95  \\[0.1ex]
Take &  21 &   30  &  44 &  95\\[0.1ex]
Typeset &  26  &  75   & 111 & 212  \\[0.1ex]
Walk &  90  & 73 & 74 & 237  \\[0.1ex]
Wash &  22 &    31 & 60 & 113   \\[0.1ex]
Whiteboard & 47  & 79    &  97 & 223 \\[0.1ex]
Write &  18  &   53  &  89 &  160 \\
\midrule
Total &  464  &   867  &  1308 & \textbf{2639} \\ \bottomrule
\end{tabular}
}
\end{table}

 \subsection{Annotation}
The dataset is provided with proper annotation for each 5-seconds video segment. It enables supervised learning in the activity recognition task. ELAN~\cite{wittenburg2006elan} tool and external camera was used to annotate video segments from the Barcelona sub-dataset. For video segments collected in Nairobi and Oxford, we followed concurrent voice annotation by having the user calling out the activity name at the start of each activity.  Finally, the preprocessing is performed to segment out each 5-seconds duration and where the audio is muted.

\subsection{Challenges}
\begin{figure*}[t]
\centering
\subfloat[]{
  \includegraphics[height=15mm,width=30mm]{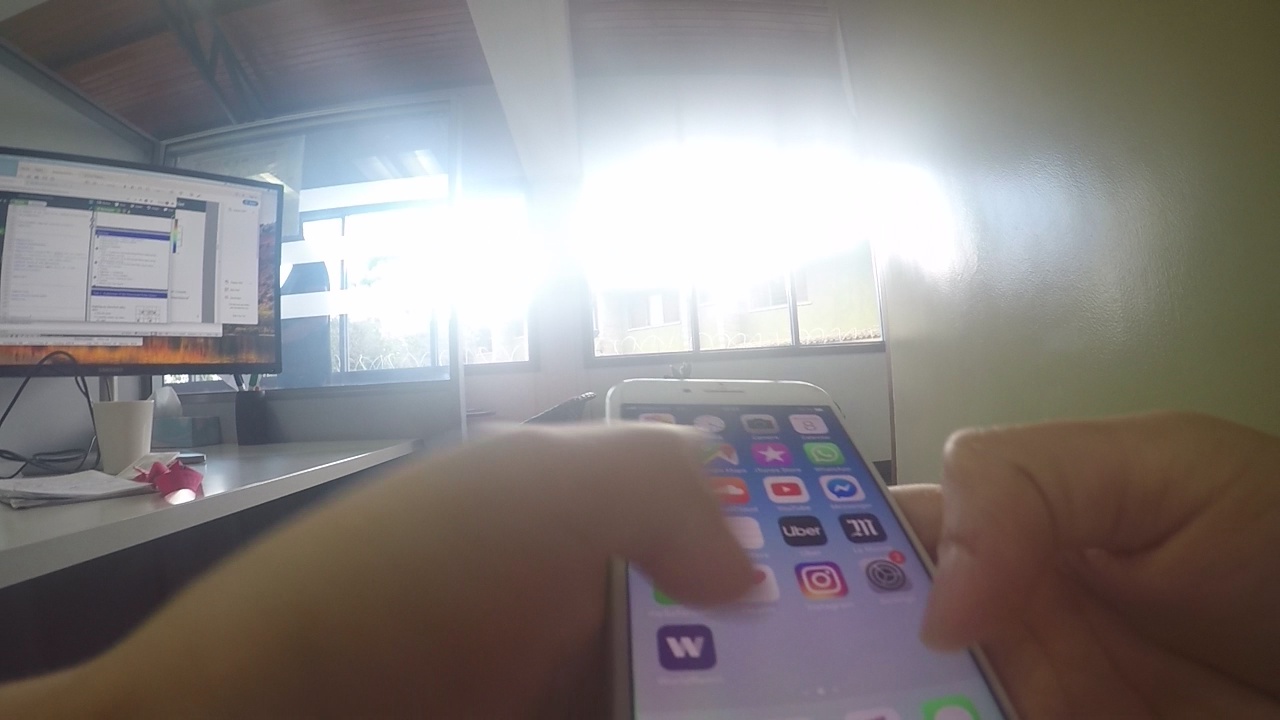}
  \label{illumch1}
}
\subfloat[]{
  \includegraphics[height=15mm,width=30mm]{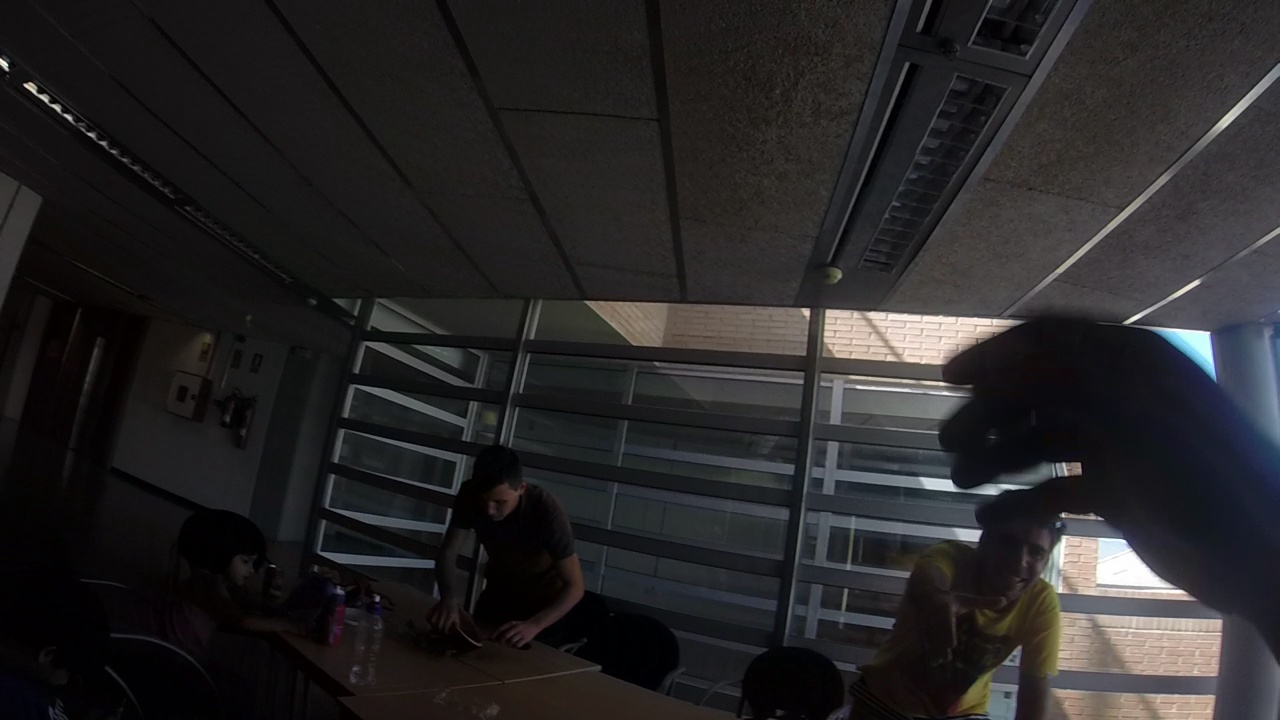}
  \label{illumch2}
}
\subfloat[]{
  \includegraphics[height=15mm,width=30mm]{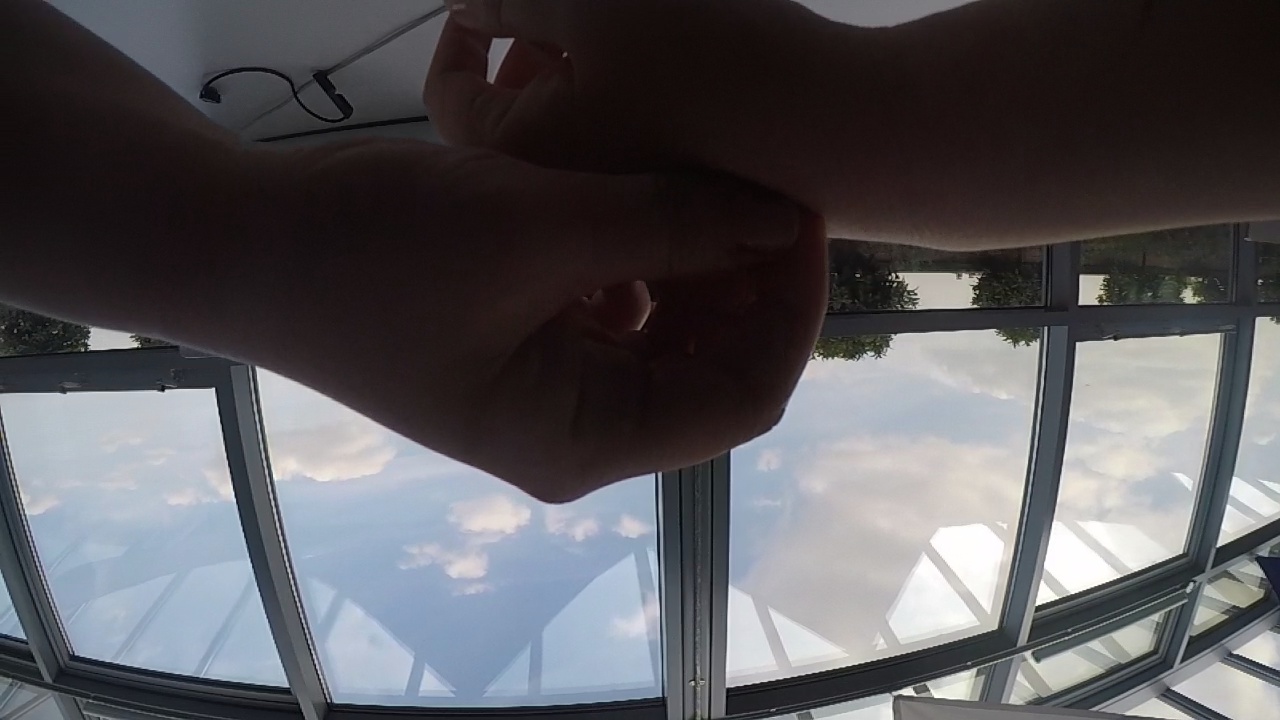}
  \label{occlus}
}
\newline
\subfloat[]{
  \includegraphics[height=15mm,width=30mm]{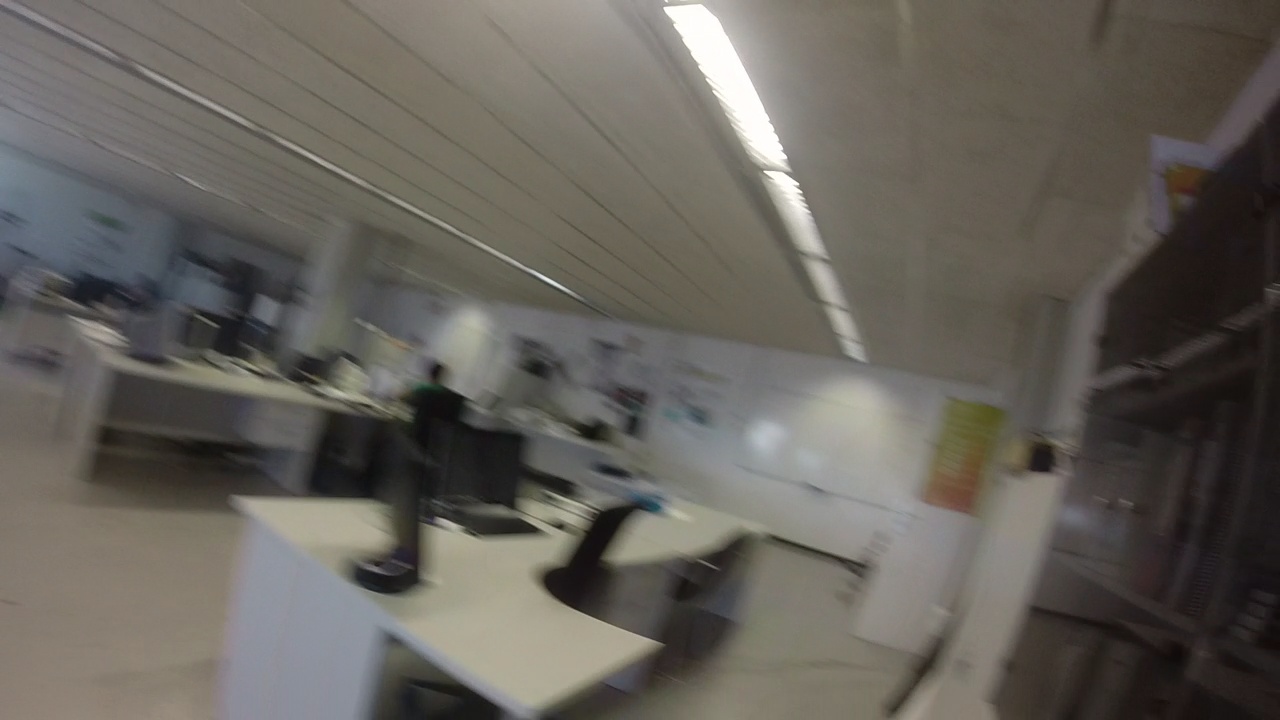}
  \label{mot_bl}
}
\subfloat[]{
  \includegraphics[height=15mm,width=30mm]{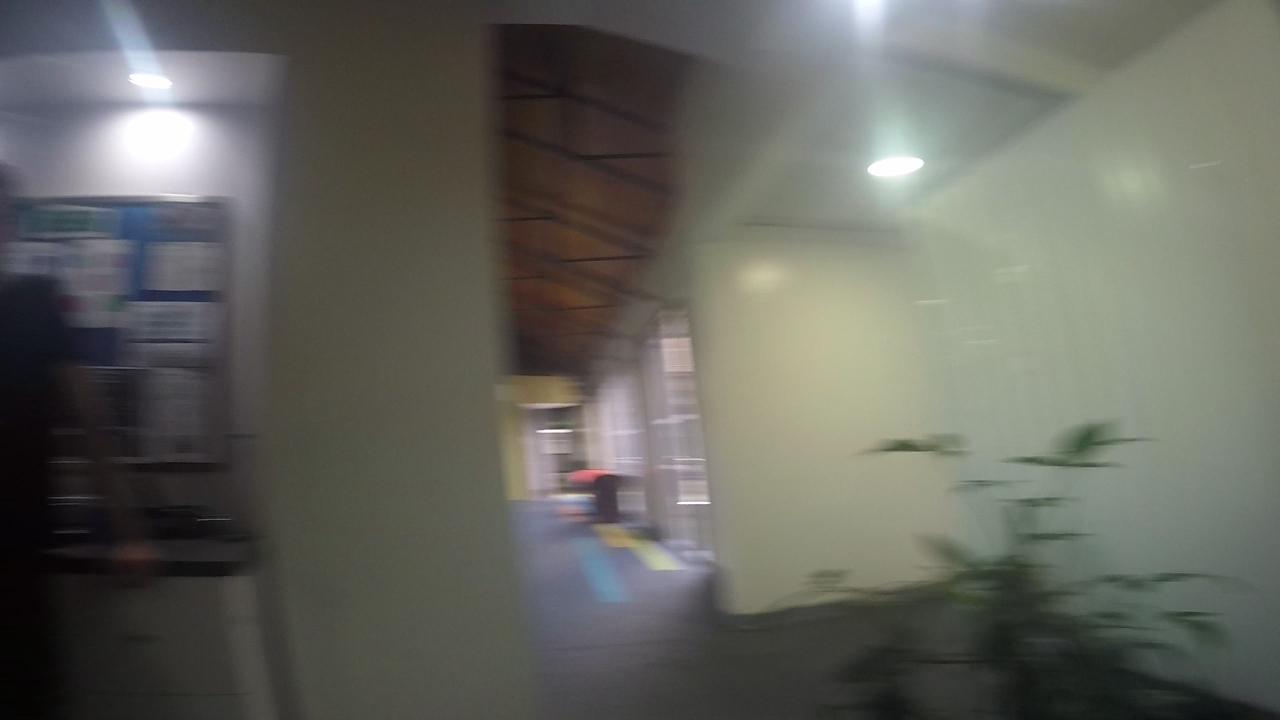}
  \label{out_mot1}
}
\subfloat[]{
  \includegraphics[height=15mm,width=30mm]{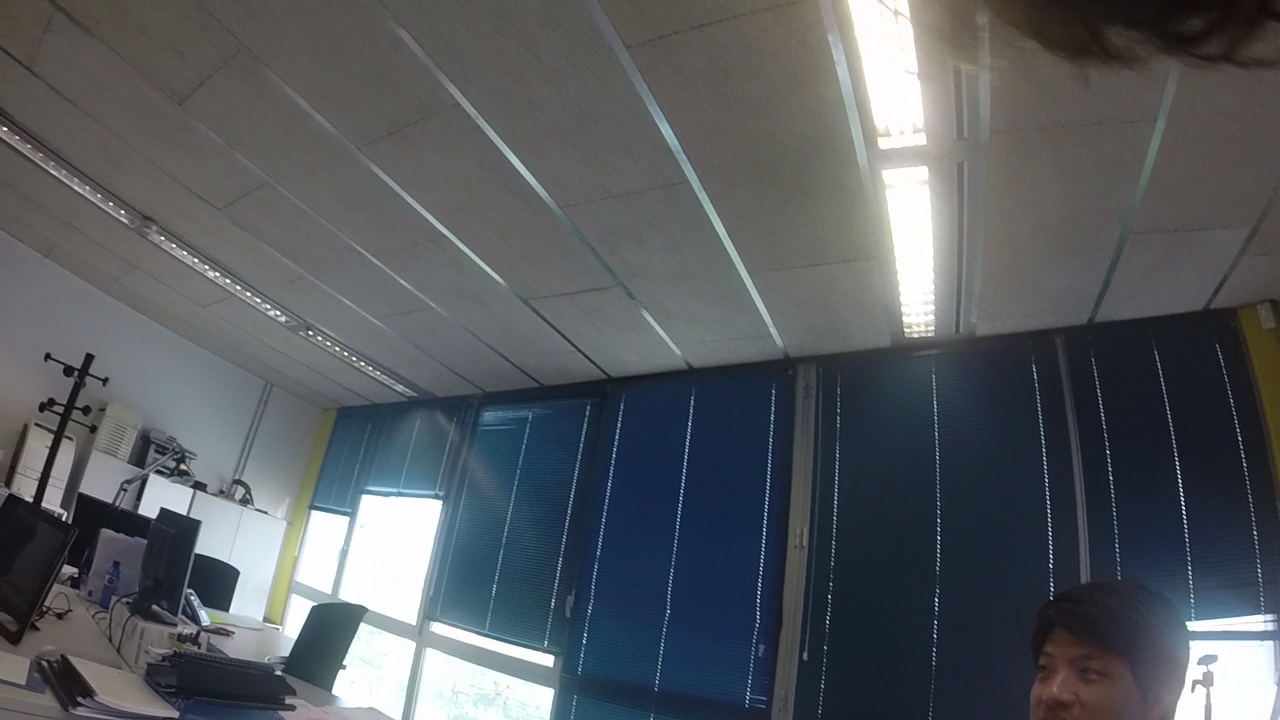}
  \label{out_mot2}
}
\newline
\subfloat[]{
  \includegraphics[height=15mm,width=30mm]{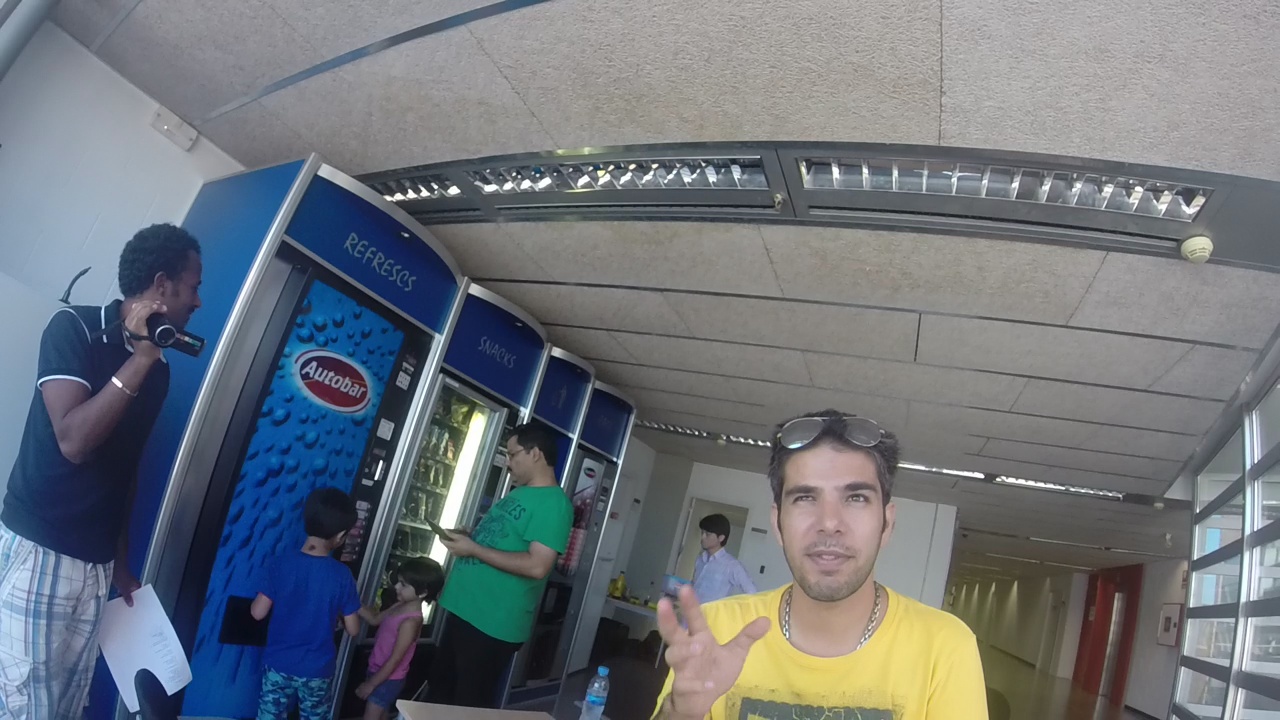}
  \label{privacy1}
}
\subfloat[]{
  \includegraphics[height=15mm,width=30mm]{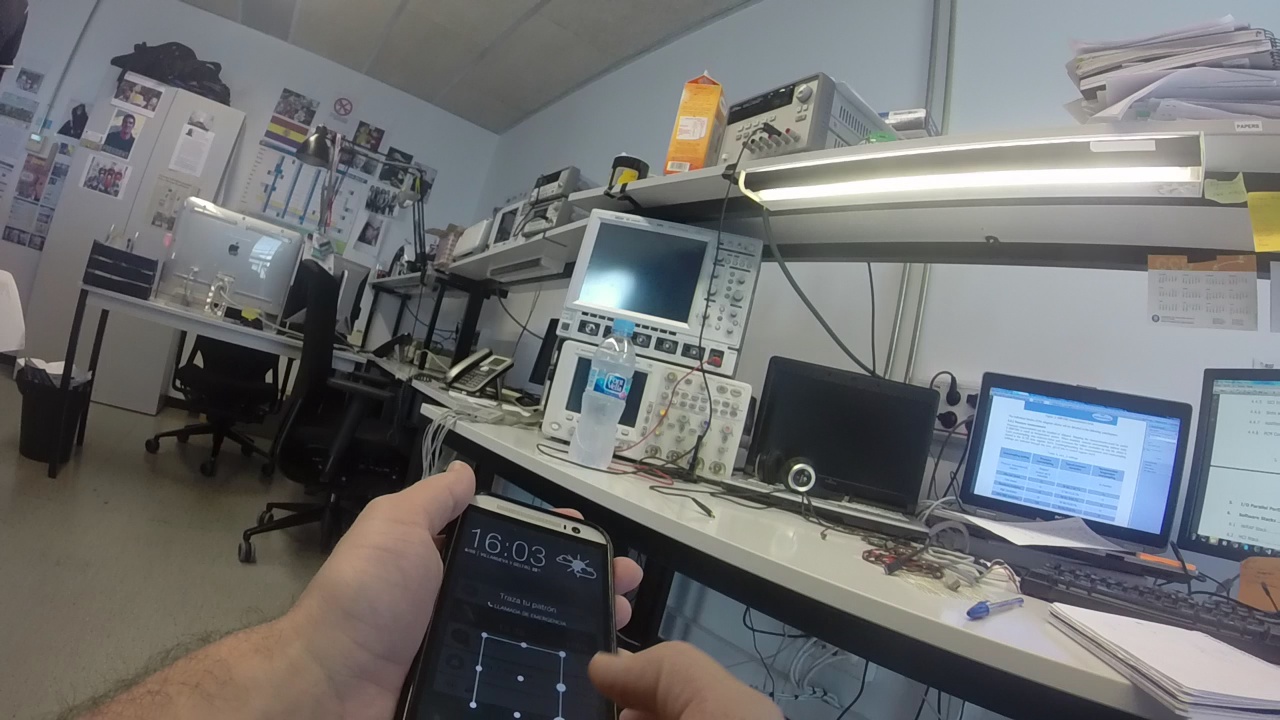}
  \label{privacy2}
}
\caption{Examples of video frames to demonstrate some of the challenges in the BON dataset: illumination changes in \emph{mobile} in (a) and \emph{shake} in (b) activities. Other challenges include occlusion  in \emph{chat}  activity in (c), some frames fade due to the motion of the subject in (d),
outlier motion happens while \emph{stapling} ends in (e),  outlier motion occurs while \emph{shaking} hand in (f), and  privacy exposed in \emph{chat}  and \emph{mobile} activities in (g) and (h), respectively, as faces of associated people and mobile password is disclosed in the frame.}
\label{challenge}
\end{figure*}

 The field of computer vision is known for  challenges, such as illumination changes, occlusion, outlier motion, and privacy, which are also reflected in the BON dataset (see Fig.~\ref{challenge}).  For example,  changes in illumination changes,  caused by unanticipated brightness variations (Fig. \ref{illumch1} and \ref{illumch2}), were observed due to the mixture of indoor and outdoor lighting. 
Occlusion, on the other hand,  can partially or completely overshadows the object of interest in the video. Example is a camera occluded by moving hands as in Fig.~\ref{occlus}. As the FPV camera is chest-mounted, movement of human body could result in outlier motion in the video scene and/or blurry frames (see Fig.~\ref{mot_bl} and \ref{out_mot1}).
Another concerning issue is privacy.
For instance, chatting with someone (Fig. \ref{privacy1}) could reveal the faces of many individuals indirectly; moreover, sensitive content could also be exposed while working on a computer or accessing  a mobile phone. Privacy-preserving in first-person videos is a growing research interest~\cite{tadesse2020privacy}, and BON could provide more data to further strengthen this research.

\begin{figure*}[t]
\centering
\subfloat[Microwave activity variations]{
  \includegraphics[height=15mm,width=30mm]{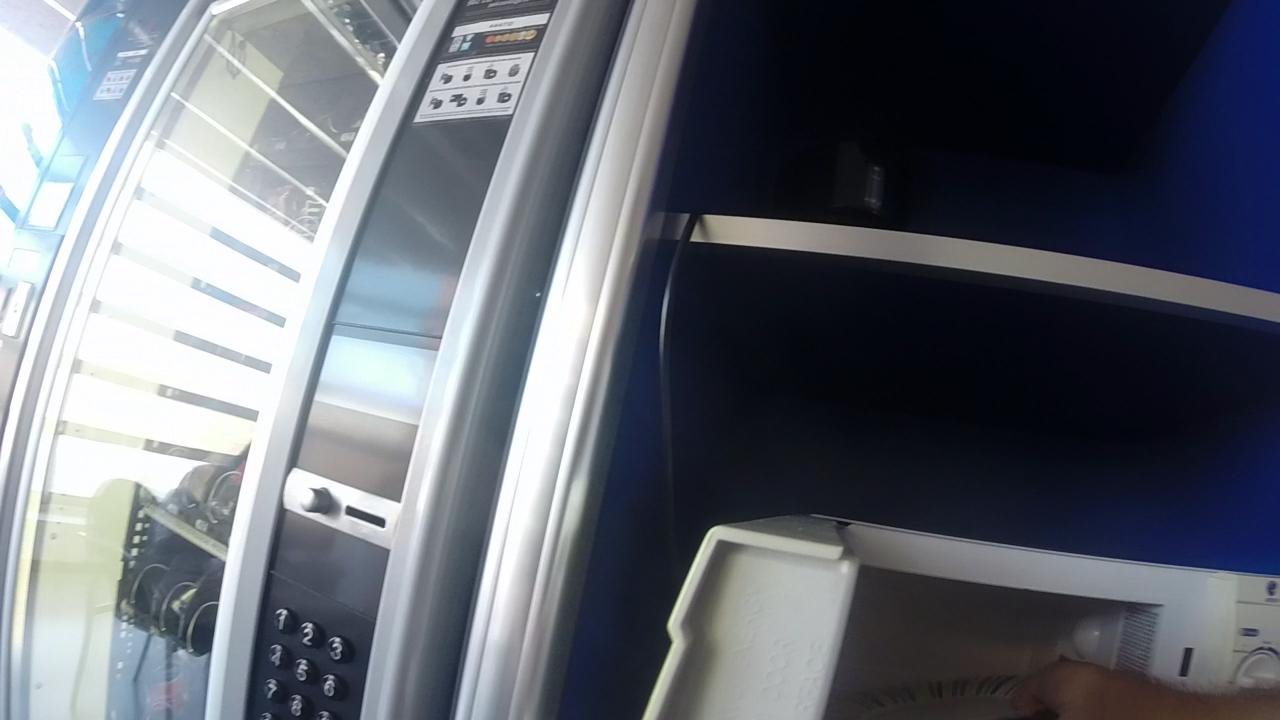}
   \includegraphics[height=15mm,width=30mm]{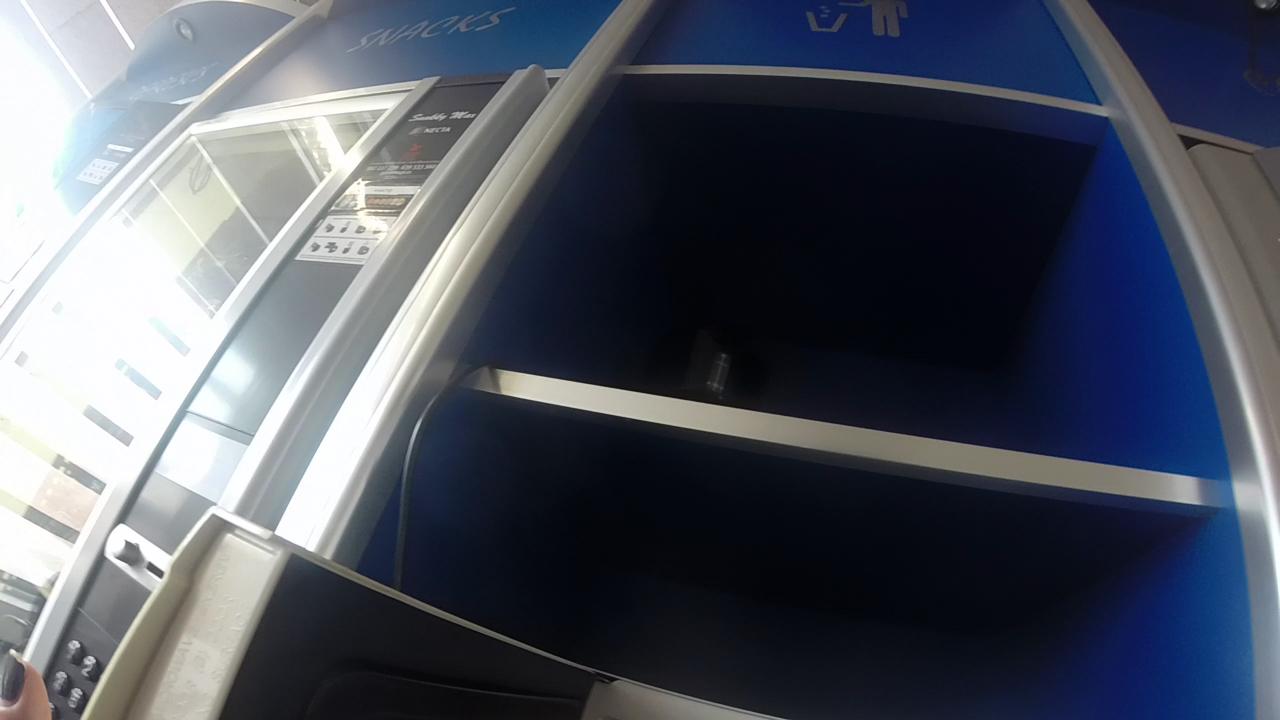}
     \includegraphics[height=15mm,width=30mm]{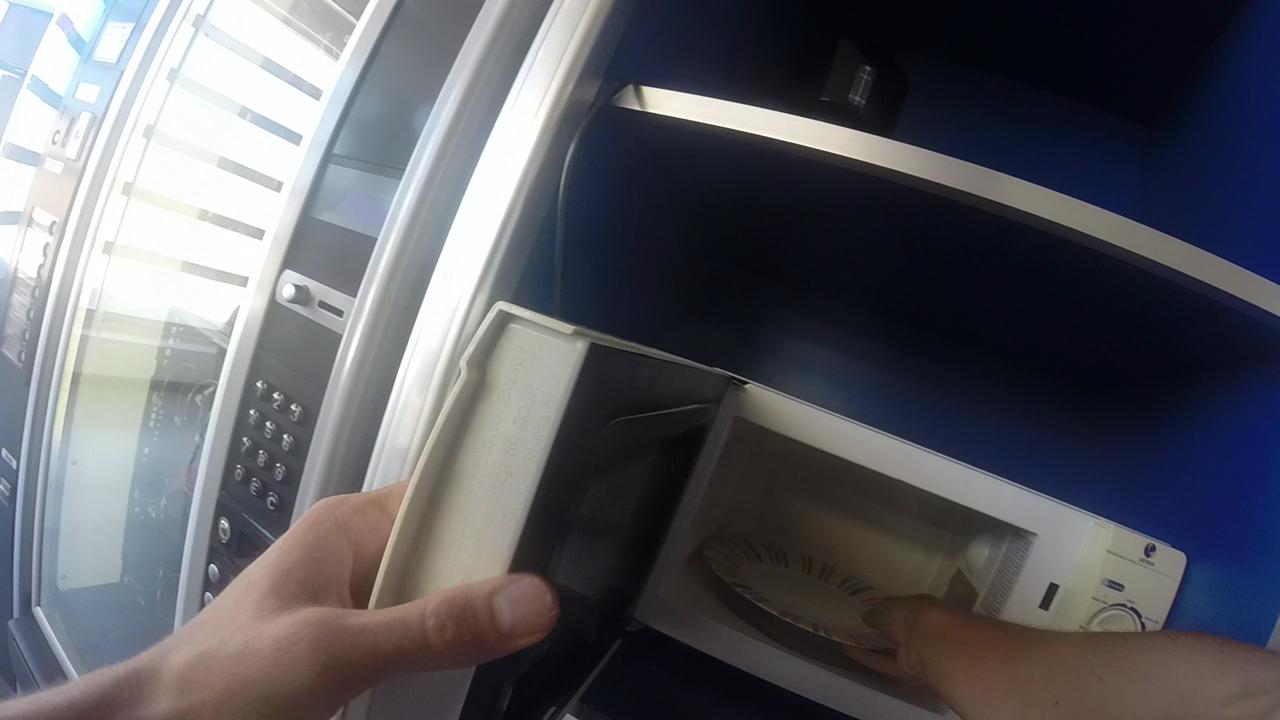}
       \includegraphics[height=15mm,width=30mm]{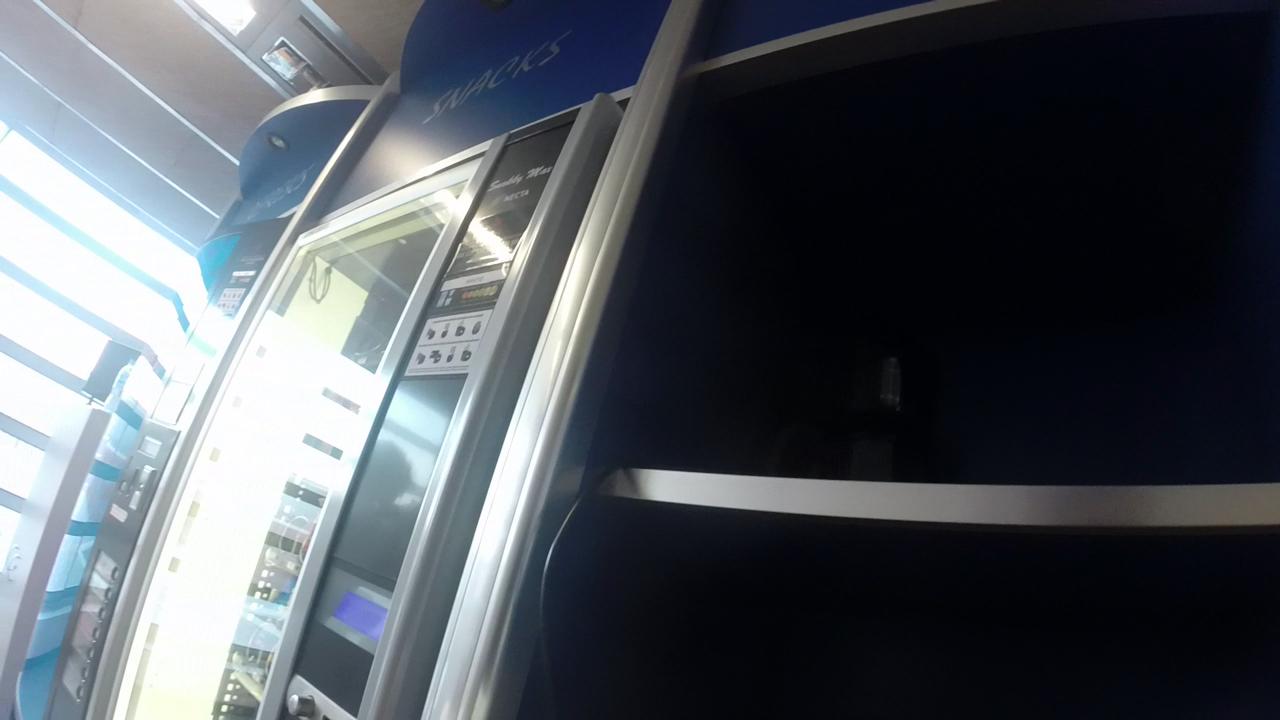}
}

\subfloat[Machine activity variations ]{
  \includegraphics[height=15mm,width=30mm]{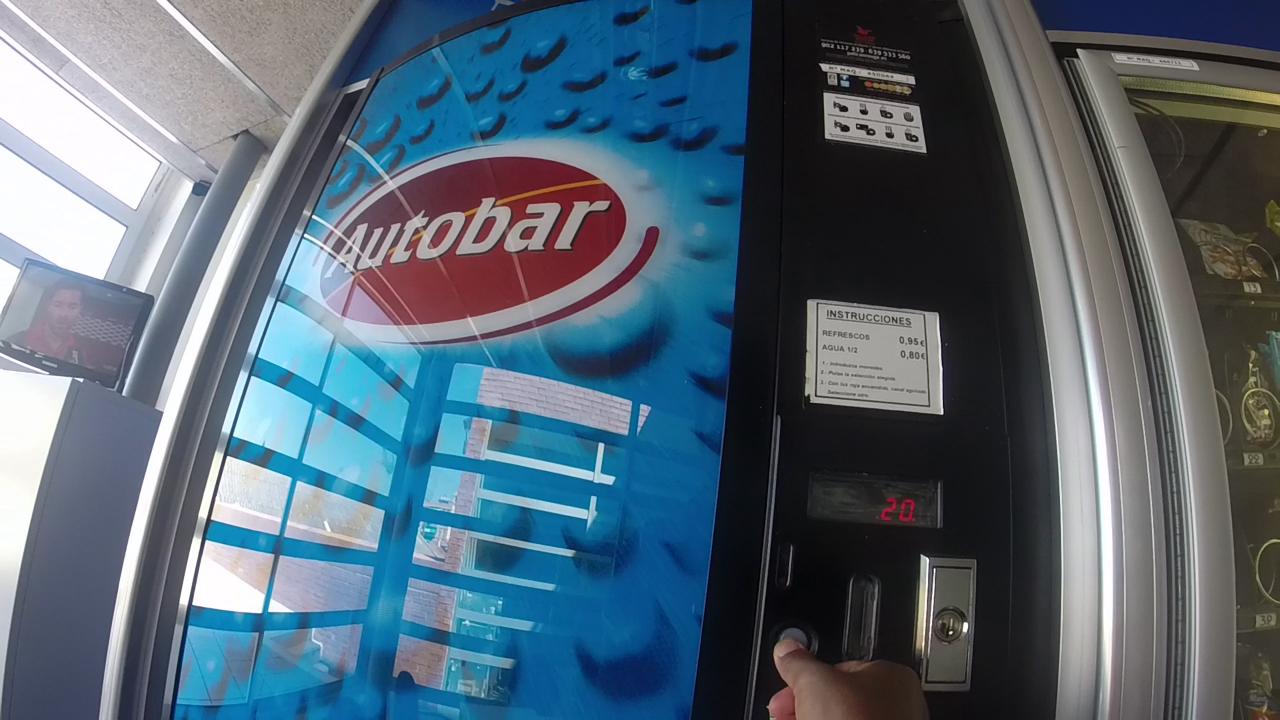}
  \includegraphics[height=15mm,width=30mm]{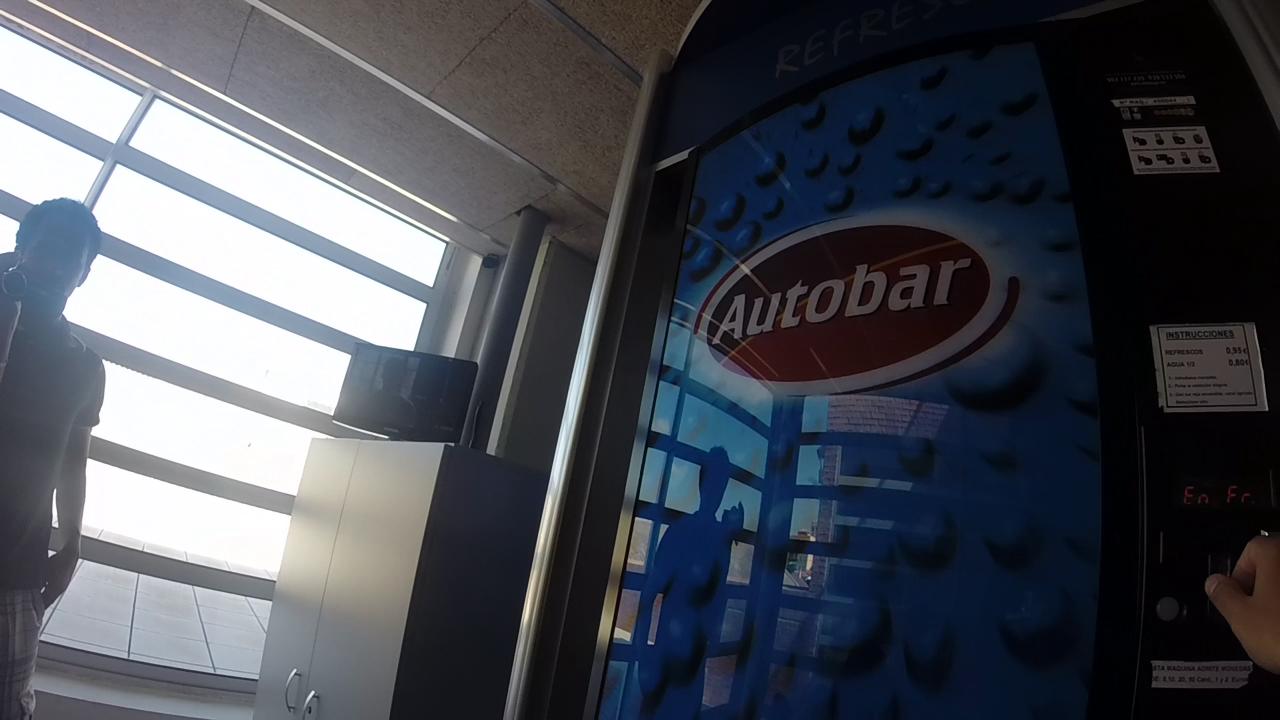}
    \includegraphics[height=15mm,width=30mm]{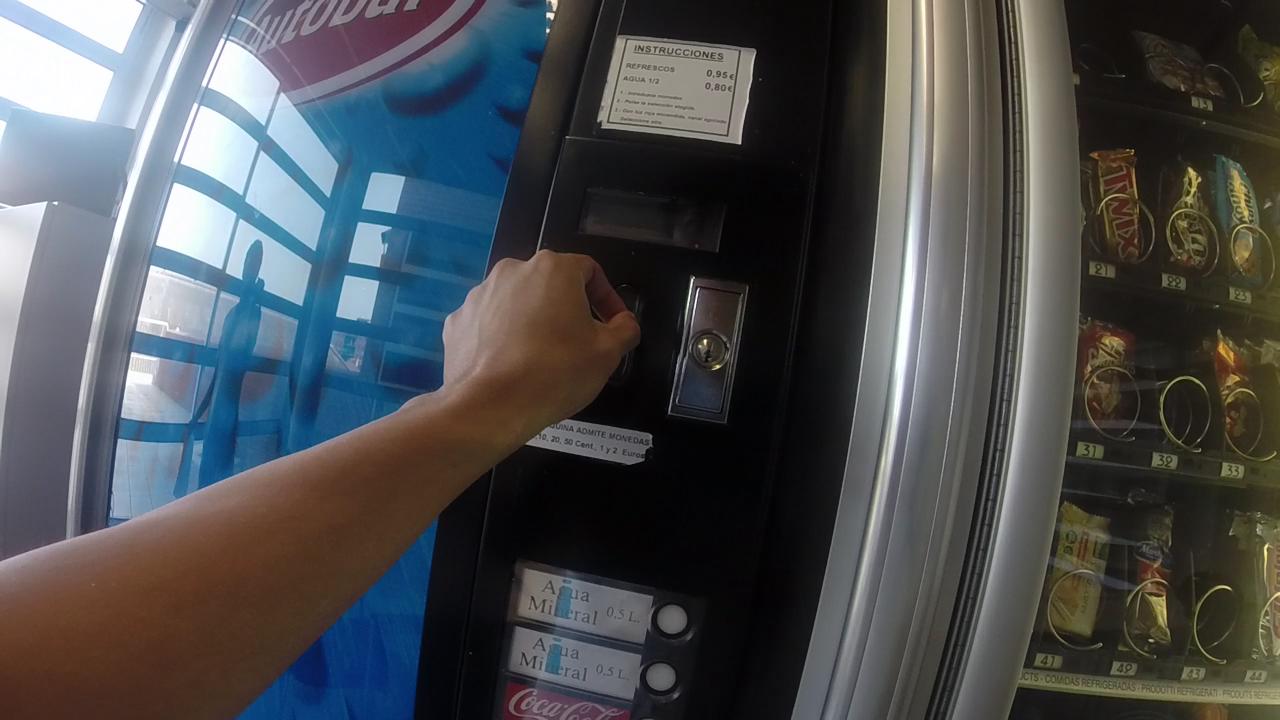}
      \includegraphics[height=15mm,width=30mm]{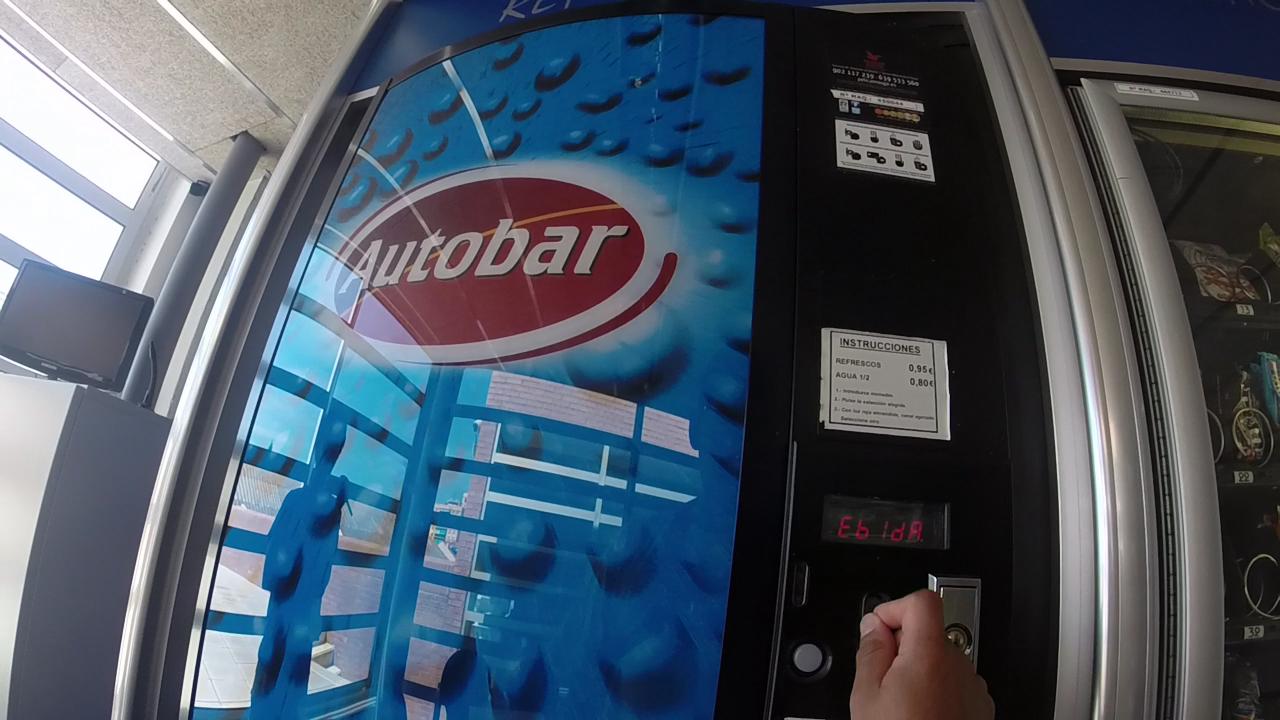}
}

\caption{Examples of intra-class variations where an activity is performed differently across subjects: (a) Microwave activity and  (b) Machine activity.}
\label{inter_subject}
\end{figure*}

\begin{figure}
\centering
\subfloat[Dry (Barcelona)]{
  \includegraphics[height=15mm,width=30mm]{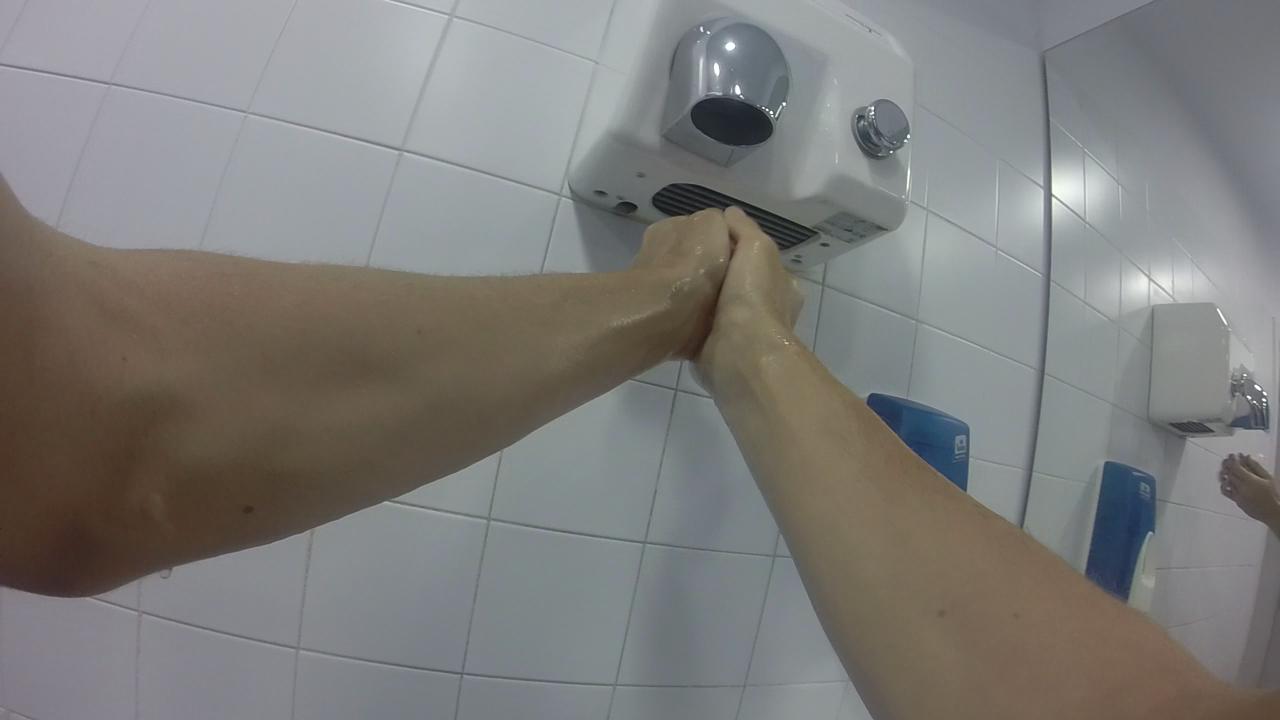}
  \label{dry_barca}
}
\subfloat[Walk (Barcelona)]{
  \includegraphics[height=15mm,width=30mm]{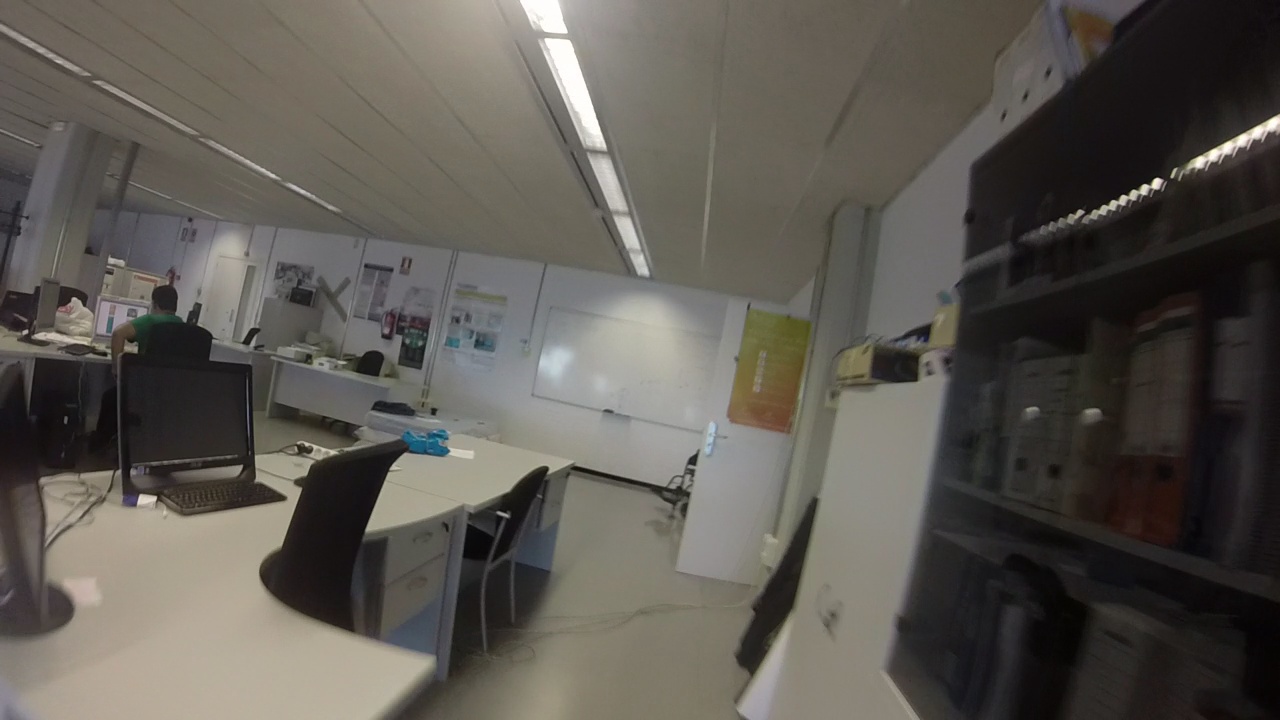}
  \label{walk_barca}
}
\\
\subfloat[Dry (Oxford)]{
  \includegraphics[height=15mm,width=30mm]{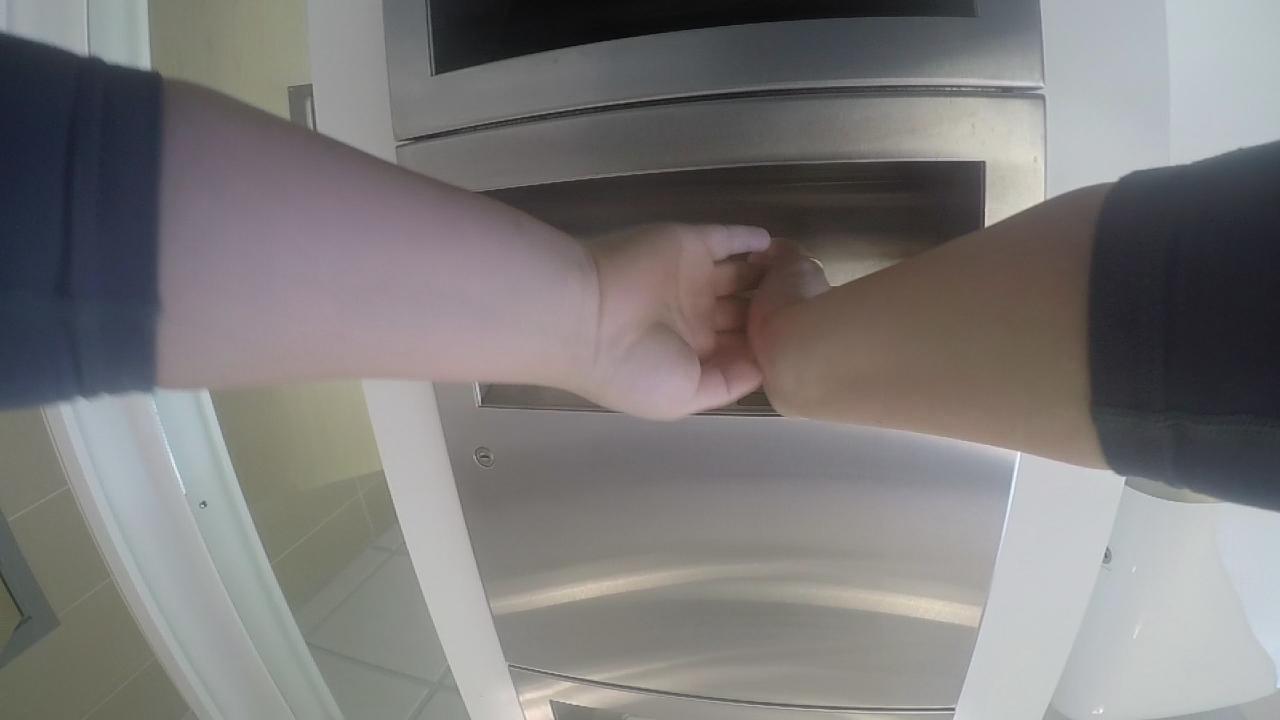}
  \label{dry_oxf}
}
\subfloat[Walk (Oxford)]{
  \includegraphics[height=15mm,width=30mm]{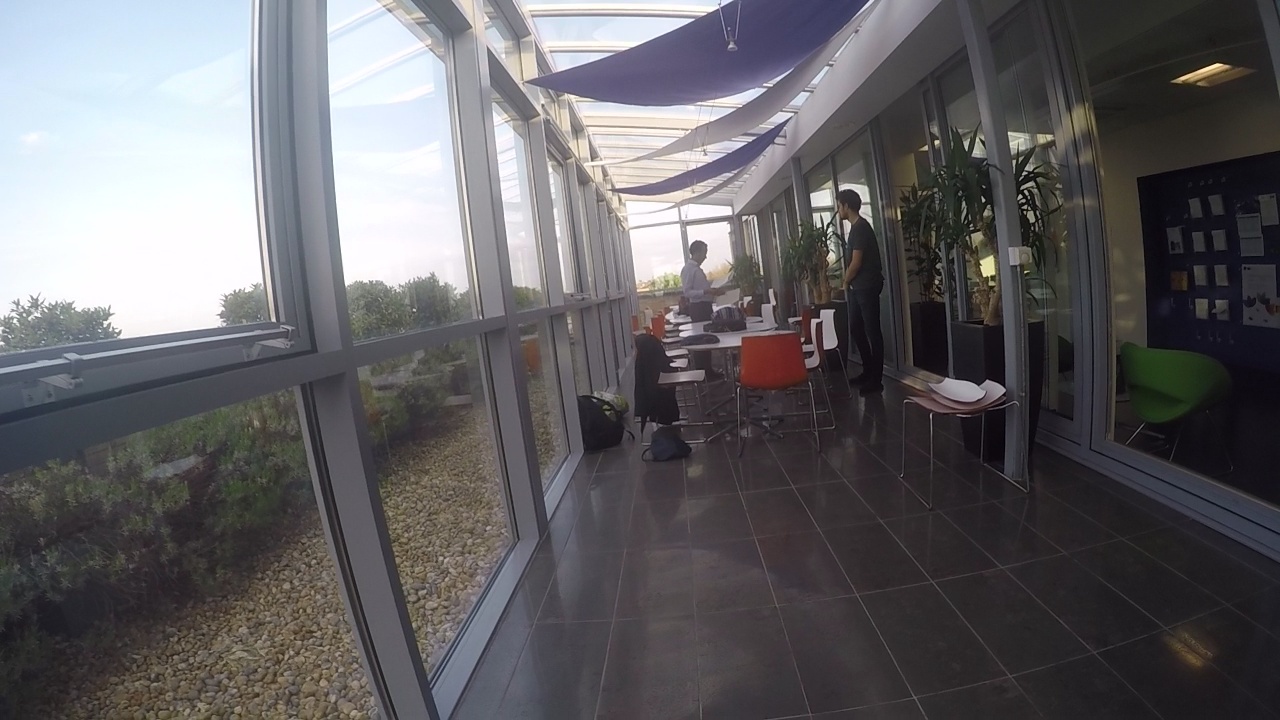}
  \label{walk_oxf}
}
\\
\subfloat[Dry (Nairobi)]{
  \includegraphics[height=15mm,width=30mm]{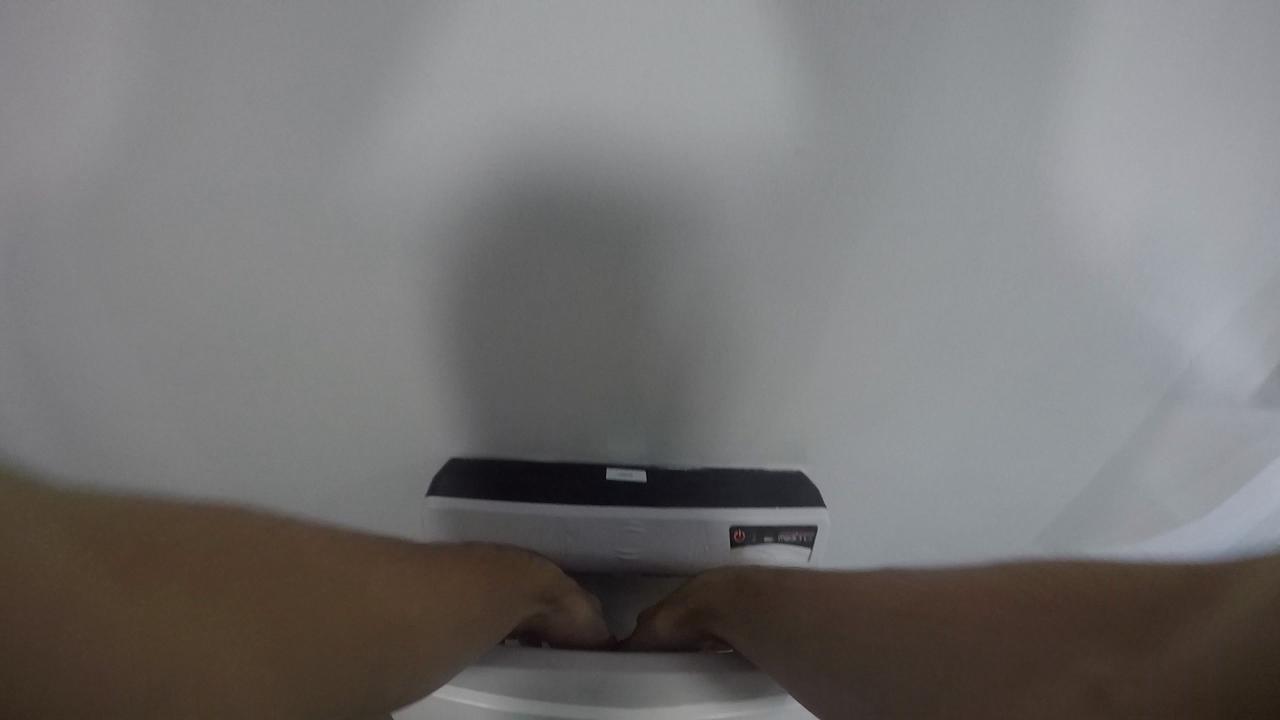}
  \label{dry_nai}
}
\subfloat[Walk (Nairobi)]{
  \includegraphics[height=15mm,width=30mm]{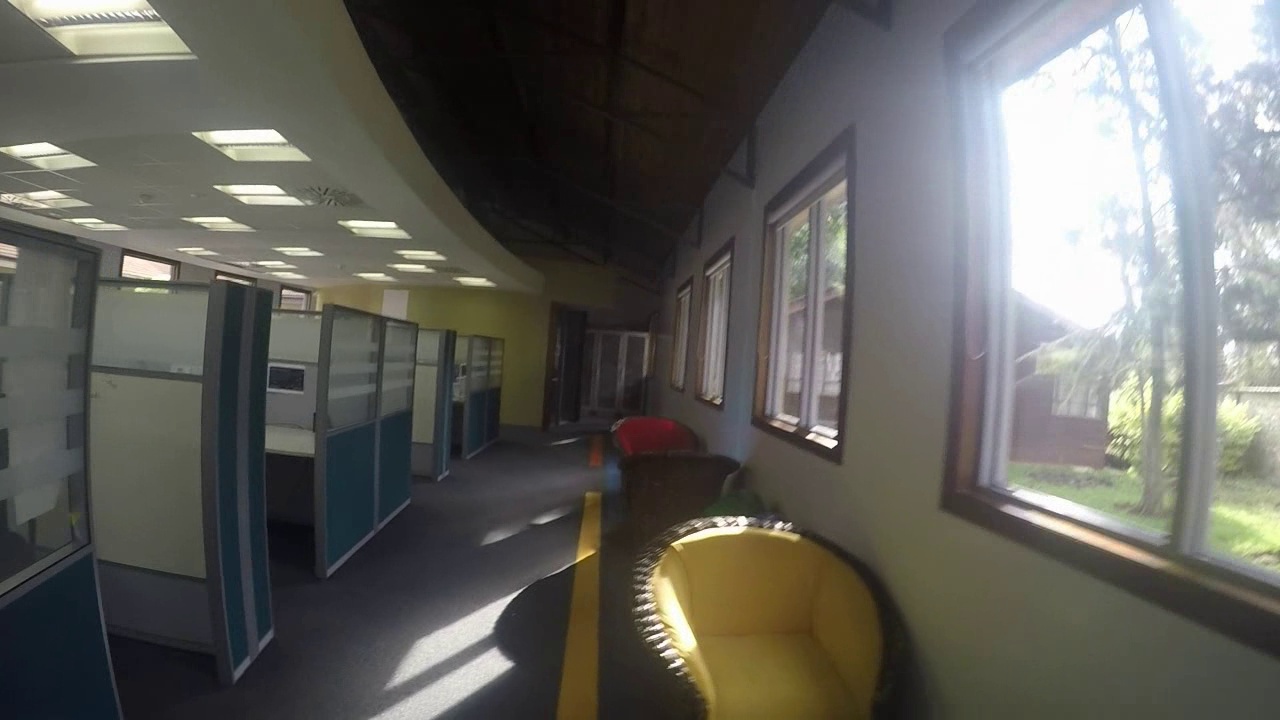}
  \label{walk_nai}
}
\caption{Variations in office settings: the first column shows different hand \textit{drying} machines and the second column shows indoor \textit{walking} environments across the three locations used for data collections: Barcelona, Oxford and Nairobi.}\label{fig:more_intra_inter}
\end{figure}

\begin{figure}[b]
\centering
\subfloat[Machine]{
  \includegraphics[height=15mm,width=30mm]{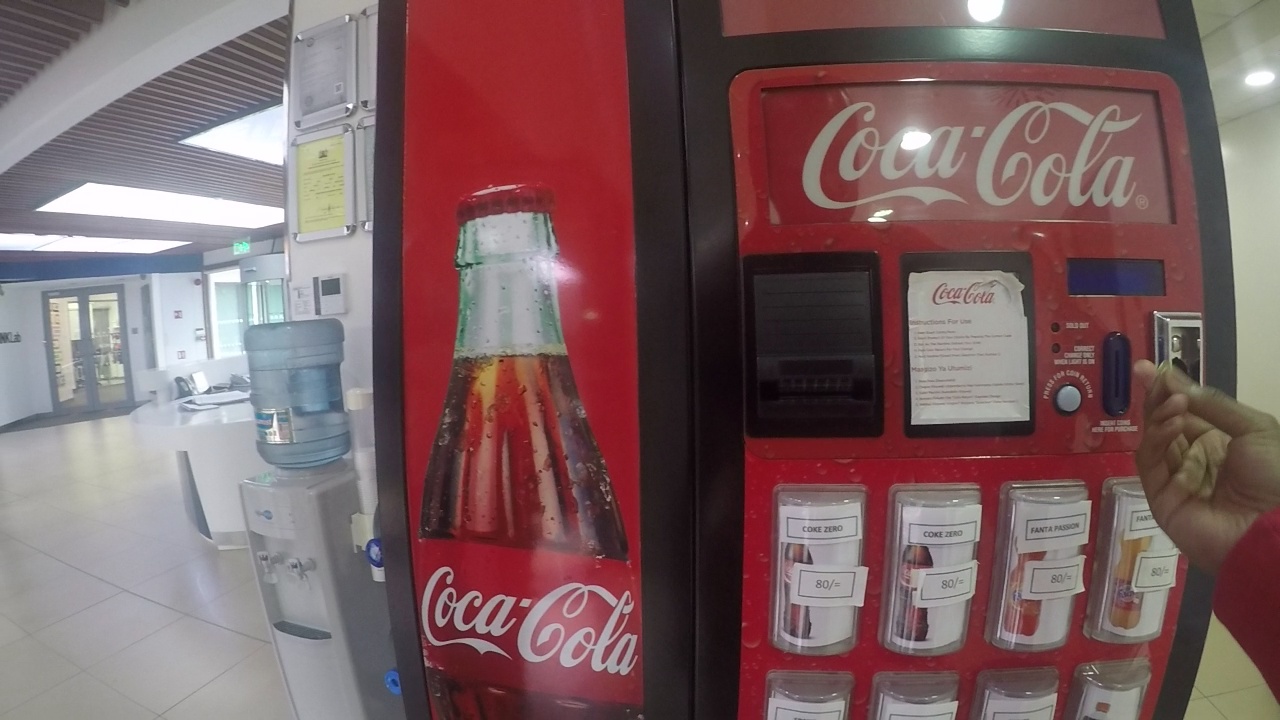}
  \label{inter_activitya}
}
\subfloat[Take]{
  \includegraphics[height=15mm,width=30mm]{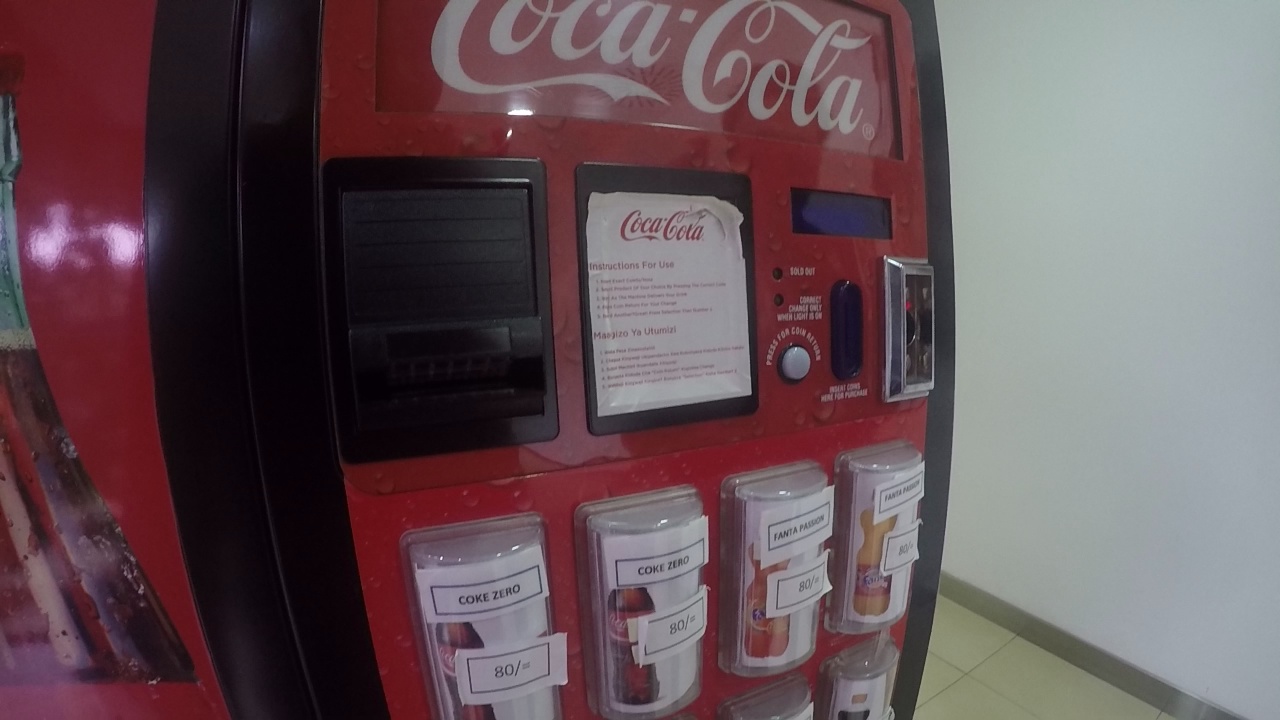}
  \label{inter_activityb}
}
\\
\subfloat[Read]{
  \includegraphics[height=15mm,width=30mm]{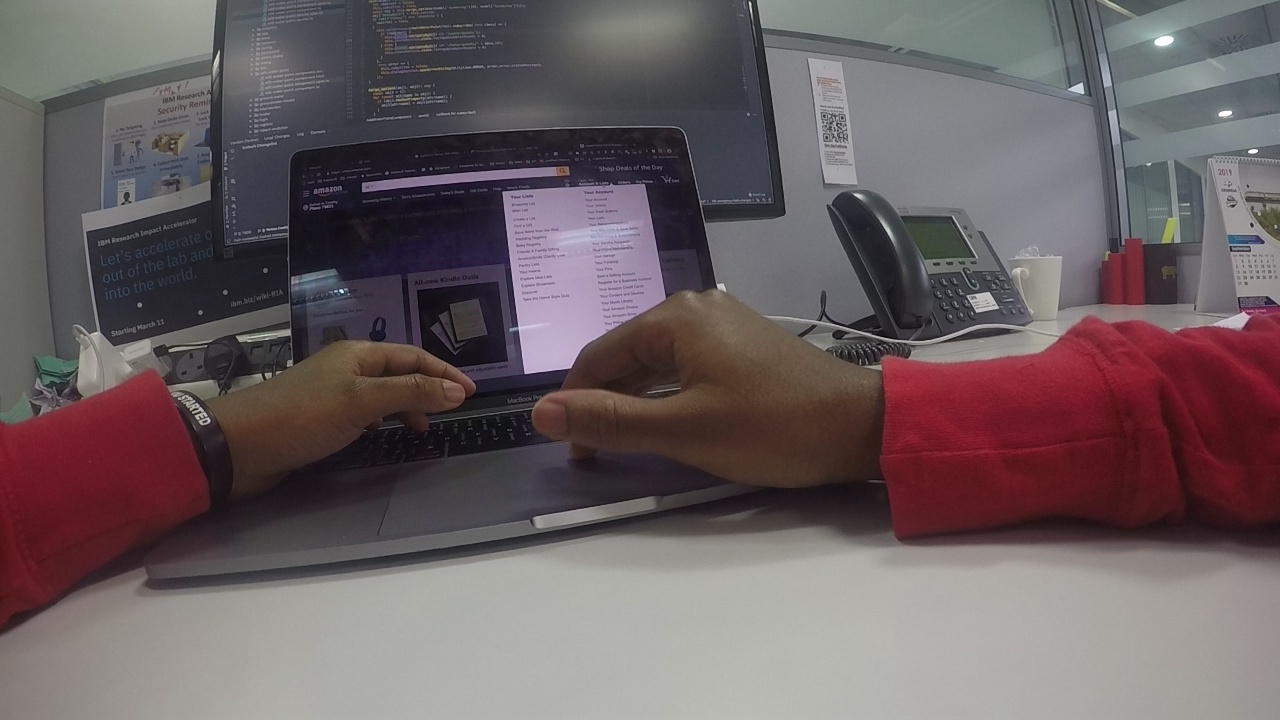}
  \label{inter_activityc}
}
\subfloat[Typeset]{
  \includegraphics[height=15mm,width=30mm]{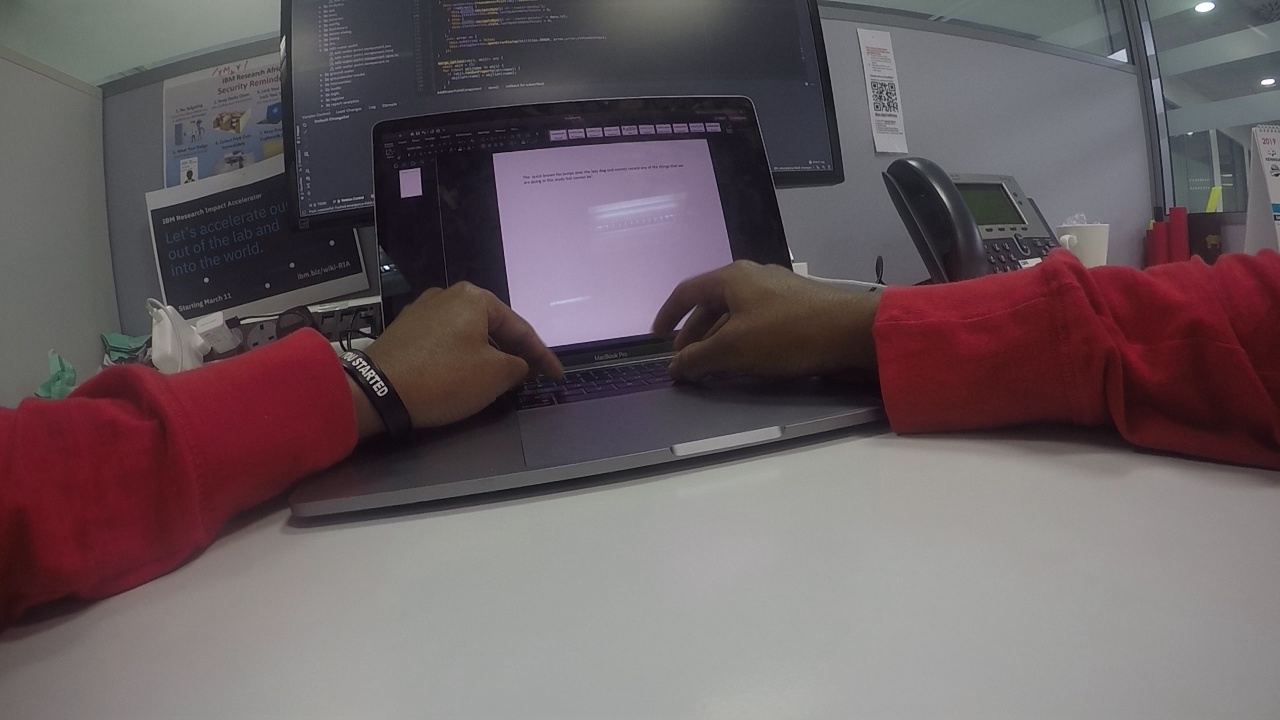}
  \label{inter_activityd}
}
\caption{Inter-activity similarity. Comparison with the first column to the second column illustrates the similarity between two different activities, top row: Machine vs. Take, bottom row: Read vs. Typeset.}
\label{inter_activity}
\end{figure}

Additional challenges evident in the BON dataset include intra-class variations  (e.g, how different subjects perform similar activity differently) and inter-class similarity (e.g., how different activities could appear similar in first-person videos) as shown in Figure~\ref{inter_subject}. Further examples are also provided in Fig.~\ref{fig:more_intra_inter} to demonstrate intra-class variations due to different office settings, e.g., different hand-drying machines.
 Examples for inter-class similarity are shown between  \emph{Read} (in Fig.~\ref{inter_activityc}) and \emph{Typeset} (in Fig.~\ref{inter_activityd}). Similarly, Fig.~\ref{inter_activitya} and Fig.~\ref{inter_activityb} for \emph{Machine} and \emph{Take}, respectively, have shown  resemblance in how the user interacts with a vending machine
To build a robust classifier, indistinguishable activities like these activities need more attention.

%

\section{Preliminary experiments}
\label{experim}
BON dataset was used  in a Video and Image Processing (VIP) Cup - a challenge organised by IEEE Signal Processing Society (SPC) at  at the IEEE International Conference on Image
Processing, 2019, Taiwan~\cite{tadesse2020privacy}.  The main task of the challenge was to recognize activities. For training,  whole video segments from Barcelona sub-dataset and the segments from the first two subjects in Oxford and Nairobi were used, whereas the remaining video segments from Oxford and Nairobi sub-datasets were used for testing. To this end, top-3 (based on average $F_1$ score) performing activity recognition frameworks were selected (see Table~\ref{baseline}) and described below. These frameworks could be utilized as baseline for future research.

Among the proposed frameworks in the IEEE VIP Cup 2019, the highest scoring baseline model used a spatial
temporal attention reasoning (STANet)~\cite{wen2019drone} for activity recognition, whereas features were extracted from both the RGB  and corresponding optical flow frames using ResNet-34~\cite{he2016deep}.
Long Short-term Memory (LSTM) network~\cite{olah2015understanding} was employed, followed by spatial attention~\cite{sudhakaran2018attention}, to encode spatio-temporal information. The second ranked framework was an end-to-end deep learning approach which employs
Inception-V3~\cite{szegedy2015going} to extract features from the video frames, which were later fed into a multilayer perceptron (MLP)~\cite{taud2018multilayer} for a final prediction.  The third ranked framework was based on ensemble of several Recurrent Neural Networks (RNNs) with framewise attention~\cite{ghosh2020privacy}. Different models pre-trained on ImageNet~\cite{deng2009imagenet} (e.g., DenseNet~\cite{huang2017densely}, Wide-ResNet~\cite{zagoruyko2016wide}) were used for extracting features from sampled video frames. The ensemble scheme was based on $F_1$ score for particular classes of different models trained on different resolutions. Additionally, to handle class imbalance in the dataset, a balanced batch  technique was employed. Categorical cross entropy is used as loss and initial layers of feature extractors were frozen during the transfer learning from ImageNet pre-trained models~\cite{ghosh2020privacy}.

\begin{table}[htbp]
\caption{Comparison of baseline systems}
\label{baseline}
\centering
\begin{tabular}{lc}
\toprule
Method & Avg F1 score \\ \midrule
STANet~\cite{wen2019drone} & \textbf{0.749}\\
MLP~\cite{taud2018multilayer} & 0.678\\
RNN Ensemble~\cite{ghosh2020privacy}  & 0.658\\ 
\bottomrule
\end{tabular}
\end{table}

\newpage
\section{Conclusion}
\label{conclu}
Egocentric vision is a growing field of research in computer vision, and unlike other sub-domains (e.g., kitchen activities~\cite{Damen2018EPICKITCHENS}), there is substantial lack of data in  understanding human activities from using egocentric vision at work places or in office environments. This paper presents a large and publicly available dataset is aimed at releasing the BON dataset - collected using a chest mounted GoPro Hero3+ camera for eighteen office activities~\cite{girmaw2021bon}. The BON dataset is extended from \cite{abebe2018first}, by including diversified office settings across three countries and  increased number of participants resulting $2639$ labeled video segments  collected from 25 participants.  Moreover, this paper describes the details of the dataset (and its video segments) stratified across locations, subjects and activities. In addition, BON was shown to include real world challenges in computer vision, such as illumination changes, occlusion, intra-class variations and inter-class similarity. As baselines for future research, we also shared three activity recognition frameworks that topped the ranking in the previous IEEE VIP Cup 2019, where the full version of BON dataset was first utilized. We hope that this publicly available dataset helps to facilitate the research efforts in the field of egocentric activity recognition from wearable cameras, particularly recognition of activities in office settings that provides a promising potential in human-computer interactions, security and product monitoring.

%
%
\bibliographystyle{IEEEtran}
\bibliography{egbib}
\end{document}